\documentclass{article}

\usepackage{microtype}
\usepackage{graphicx}
\usepackage{subcaption}
\usepackage{booktabs} % for professional tables
\usepackage[ruled,vlined]{algorithm2e} % for \KwIn and other algorithm2e commands

\usepackage{wrapfig} % for wraptable environment

\usepackage{hyperref}
\usepackage{placeins}

\usepackage[preprint]{icml2026}

\usepackage{amsmath}
\usepackage{amssymb}
\usepackage{mathtools}
\usepackage{amsthm}
\usepackage{multirow}

\usepackage[capitalize,noabbrev]{cleveref}

\theoremstyle{plain}

\theoremstyle{definition}

\theoremstyle{remark}

\usepackage[textsize=small, textwidth=2cm]{todonotes}
\usepackage{xcolor}
\icmltitlerunning{Towards Universal Representation-Based Process Control}

\begin{document}

\twocolumn[
%\icmltitle{From Stationarity to Reference-Relative Hypothesis Testing in Time Series}
\icmltitle{Towards Universal Representation-Based Process Control}
% It is OKAY to include author information, even for blind
% submissions: the style file will automatically remove it for you
% unless you've provided the [accepted] option to the icml2025
% package.

% List of affiliations: The first argument should be a (short)
% identifier you will use later to specify author affiliations
% Academic affiliations should list Department, University, City, Region, Country
% Industry affiliations should list Company, City, Region, Country

% You can specify symbols, otherwise they are numbered in order.
% Ideally, you should not use this facility. Affiliations will be numbered
% in order of appearance and this is the preferred way.
% \icmlsetsymbol{equal}{*}

\begin{icmlauthorlist}
\icmlauthor{Jinmyeong Choi}{cmu,snu}
\icmlauthor{Taesup Kim}{snu}
\icmlauthor{Artur Dubrawski}{cmu}

\end{icmlauthorlist}

\icmlaffiliation{snu}{Graduate School of Data Science, Seoul National University, Seoul, Korea}
\icmlaffiliation{cmu}{Auton Lab, Robotics Institute, Carnegie Mellon University, Pittsburgh, USA}

\icmlcorrespondingauthor{Taesup Kim}{taesup.kim@snu.ac.kr}
\icmlcorrespondingauthor{Artur Dubrawski}{awd@andrew.cmu.edu}

% You may provide any keywords that you
% find helpful for describing your paper; these are used to populate
% the "keywords" metadata in the PDF but will not be shown in the document
\icmlkeywords{Time Series Foundation Models, Nonparametric Hypothesis Testing, Time Series Monitoring}

\vskip 0.3in
]

% this must go after the closing bracket ] following \twocolumn[ ...

% This command actually creates the footnote in the first column
% listing the affiliations and the copyright notice.
% The command takes one argument, which is text to display at the start of the footnote.
% The \icmlEqualContribution command is standard text for equal contribution.
% Remove it (just {}) if you do not need this facility.

\printAffiliationsAndNotice{}  % leave blank if no need to mention equal contribution
% \printAffiliationsAndNotice{\icmlEqualContribution} % otherwise use the standard text.

\begin{abstract}

Many temporal process learning and monitoring pipelines operate in local windows, making window-level decisions unavoidable in practice. In such settings, classical statistical tests can be applied to individual windows, but they typically evaluate predefined parametric hypotheses—such as unit-root or moment-based conditions—thereby limiting flexibility when reference behavior is defined empirically from task- or domain-specific data. In this work, we view window-level monitoring as a process control problem and reformulate it as reference-based hypothesis testing, where the null hypothesis is specified by an empirical reference distribution rather than a fixed parametric model. We operationalize this perspective through a representation-based, nonparametric framework that combines pretrained time series encoders, kernel density estimation, and conformal calibration, yielding finite-sample valid inference in learned representation space. Classical notions such as stationarity and cyclostationarity arise as natural instantiations of empirical reference sets within this framework. Through experiments, we demonstrate sensitivity to window-level distributional deviations while maintaining well-calibrated inference under stable reference regimes, highlighting the applicability of the proposed approach to a broad class of time series process control and monitoring tasks.

\end{abstract}
\section{Introduction}

Modern time series systems increasingly operate under streaming and online settings, where decisions must be made on short, rolling windows of data rather than on complete sequences\cite{li2025online, gama, online}. Examples include monitoring data quality in deployed models, detecting distributional drift in streaming inputs, and validating whether a system continues to operate under expected conditions~\cite{Fretheim748}. In such settings, window-level decisions are not a design choice but a structural necessity: systems must continuously assess their current state of operation based on limited, local observations.

This setting naturally gives rise to a class of problems traditionally referred to as \emph{process control}. Unlike classification or forecasting tasks, window-level process control poses a distinct question: \emph{is the system operating consistently with an empirical reference regime?} This formulation fundamentally distinguishes process control from traditional anomaly detection~\cite{burger2025distributionfreeprocessmonitoringconformal, indus_pc, Benneyan458}.

While anomaly detection typically targets rare outliers or extreme observations relative to a global notion of normality, process control is inherently \emph{reference-relative} and focuses on \emph{regime consistency}. A system may continue to produce locally plausible or even frequent patterns, yet gradually drift away from its stable operating conditions—through variance shifts, trend emergence, or disruptions of cyclostationary structure. Such deviations are often inconspicuous to rarity-based or pointwise thresholding methods, but are critical from an operational perspective, where the goal is not to flag unusual samples, but to detect loss of consistency with a designated reference regime.

Many existing tools for time series diagnostics, however, implicitly assume a predefined notion of normality. Classical stationarity tests such as the Augmented Dickey--Fuller (ADF) and KPSS tests exemplify this perspective. While theoretically well-founded for testing parametric hypotheses like unit roots or deterministic trends~\cite{ADF, kpss, pptest}, they are designed to answer whether a series satisfies a specific statistical definition, rather than assessing consistency with a learned baseline. Consequently, when applied to short windows, these tests often suffer from low power, unstable calibration, or insensitivity to structural deviations that preserve local stability\cite{testforunitroot, cambridge, sjosten2022comparative}.

More fundamentally, the assumption that stationarity is the sole notion of normal operation is frequently violated. Many real-world systems exhibit \emph{cyclostationary} behavior---structured temporal variations that are stable and repeatable yet violate weak stationarity by definition. From a process control perspective, such periodic variation is not abnormal but rather the expected mode of operation\cite{9514488, cyclo_a}. Treating all non-stationary signals as deviations conflates fundamentally different notions of abnormality.

Crucially, our goal is not to test whether a signal is strictly stationary or cyclostationary, but to treat both as valid stable regimes. We ask whether a given window remains \emph{consistent} with a reference regime defined by such processes. To this end, we propose a \emph{representation-based, nonparametric framework} for reference-relative process control. By embedding local windows into a learned representation space using a pretrained encoder and applying conformal calibration, we achieve finite-sample-valid inference without relying on fixed parametric assumptions. This approach enables a unified treatment of diverse temporal behaviors, applying equally to stationary and cyclostationary reference sets.
\vspace{-0.5em}

\paragraph{Contributions.} This paper makes the following contributions:
\vspace{-1.0em}
\begin{itemize}
    \item We formulate window-level time series monitoring as a \emph{reference-relative process control problem}, where normality is defined empirically by a reference regime rather than fixed parametric hypotheses.
    \item We propose a \emph{representation-based, nonparametric framework} with conformal calibration that yields decision-ready, finite-sample-valid assessments of regime consistency.
    \item We demonstrate that the framework naturally accommodates \emph{cyclostationary processes as stable operating regimes}, addressing a fundamental limitation of classical stationarity-based diagnostics.
    \item Through extensive experiments, we show robustness to structural deviations and illustrate the framework's potential for diagnostic attribution beyond binary detection.
\end{itemize}
\vspace{-0.5em}

\section{Related Work}

\paragraph{Process Control and Statistical Monitoring.}
Statistical process control (SPC) has a long history in quality control and industrial monitoring, where the primary objective is to determine whether a system remains in an \emph{in-control} state relative to a reference distribution estimated from historical data \cite{montgomery2020introduction, burger2025distributionfreeprocessmonitoringconformal}. Classical SPC methods typically rely on low-dimensional summary statistics, such as means or variances, combined with parametric assumptions and fixed control limits, as exemplified by Shewhart charts, CUSUM, and EWMA procedures \cite{Page1954CONTINUOUSIS, Roberts2000ControlCT, math11163444}. While these methods provide principled control limits under strong modeling assumptions,
they are difficult to apply directly to high-dimensional representations or heterogeneous windowed signals,
where defining appropriate summary statistics is nontrivial.

\paragraph{Stationarity and Cyclostationarity Testing.}
A large body of literature is devoted to testing stationarity and related properties in time series. Classical tests such as the Augmented Dickey--Fuller (ADF) and KPSS tests assess predefined parametric hypotheses, for example the presence of a unit root or a deterministic trend \cite{ADF, kpss}. Closely related approaches, such as the Phillips--Perron (PP) test, address serial correlation and heteroskedasticity through nonparametric long-run variance corrections \cite{pptest}, while remaining focused on testing specific unit-root hypotheses. 

Extensions to cyclostationarity focus on detecting periodic structure in second-order statistics, often through spectral or cyclic correlation analysis \cite{Gardner_1986, Napolitano_2016, DG_TEST}. These methods are theoretically well-grounded and effective for identifying specific forms of (cyclo)stationarity. However, they are designed to answer a fundamentally different question from ours: whether a time series satisfies a particular statistical definition. 
\vspace{-0.5em}
\paragraph{Change-Point Detection and Structural Breaks.}
Change-point detection (CPD) methods aim to identify the time indices at which the statistical properties of a sequence change \cite{truong}. This includes both classical approaches based on likelihood ratios or cumulative sums \cite{basseville} and more recent methods leveraging kernel statistics, spectral features, or deep representations \cite{Aminikhanghahi2016ASO}. These methods are complementary to ours, but address a different objective:
localizing changes over time rather than assessing window-level consistency
with respect to a fixed reference regime.
\vspace{-0.5em}
\paragraph{Anomaly Detection and Distribution Shift.}
Anomaly detection methods seek to identify rare or unusual observations that deviate from a learned notion of normality \cite{chandola}. In time series settings, this includes approaches based on reconstruction error, prediction residuals, or density estimation in learned feature spaces \cite{hundman2018detecting}. Relatedly, distribution shift and out-of-distribution (OOD) detection focus on identifying global changes between training and test distributions \cite{quionero, ood_ts}. While these lines of work are highly relevant to reliability and safety, they typically define abnormality in terms of rarity or population-level deviation. In contrast, process monitoring often requires detecting deviations
that may be common in a global sense but unacceptable relative to a specific reference context.
This motivates reference-relative diagnostics with explicit false-alarm control,
rather than population-level anomaly scoring.

\paragraph{Representation Learning for Time Series Monitoring.}
Recent advances in representation learning have enabled powerful pretrained models for time series analysis, particularly in forecasting and self-supervised learning settings.\cite{chronos1, chronos2, moment, moirai, moirai_moe, TiRex} Several works have explored the use of learned representations for downstream tasks such as anomaly detection or change-point detection \cite{TNC, survey_tsfm, WEHNER2025111395, ood_ts}. However, pretrained representations are typically high-dimensional and encode a wide range of information\cite{wilinski2025exploring},
much of which is irrelevant to statistical stability.
This raises the question of how to orient representation spaces toward
directions that are most sensitive to distributional deviation,
which is a central focus of our methodology.
\vspace{-0.5em}
\section{Methodology}

\subsection{Problem Formulation: Reference-Relative Process Control}

We consider a window-level monitoring problem in which decisions are made on
local segments extracted from a potentially long or streaming time series.
Each window constitutes an independent unit of analysis, and the objective is to determine whether a given window remains consistent with an expected mode of operation.

Let $\mathcal{R} = \{x_i\}_{i=1}^N$ denote a collection of reference windows representing normal operation.
The reference set is defined empirically, for example from historical data, and is assumed to capture the distributional variability of the operating
regime of interest.
Given a test window $x_{\text{test}}$, the task is to assess whether
$x_{\text{test}}$ is distributionally consistent with the reference set
$\mathcal{R}$.

Crucially, no parametric assumptions are imposed on the reference regime.
The windows in $\mathcal{R}$ may arise from stationary processes,
cyclostationary processes, or other temporally structured but stable behaviors.
Rather than testing a predefined statistical hypothesis, the decision is made
relative to the empirical distribution induced by the reference set.

This formulation differs from change-point detection and global sequence-level
analysis.
Windows are evaluated independently, without attempting to localize when a
change occurs or to exploit temporal continuity between windows.
The output is a window-level decision intended for operational use, enabling
in-control versus out-of-control judgments with respect to a specified
reference regime.

By changing the construction of the reference set $\mathcal{R}$, the same
framework can be applied to different notions of normality without modifying
the underlying methodology.
In the following sections, we describe a representation-based instantiation of
this reference-relative process control formulation.

\subsection{Representation-Based Consistency Testing}

Directly assessing distributional consistency in raw time series space is challenging,
as short windows often exhibit substantial variability due to noise, phase effects,
and local fluctuations.
This issue is exacerbated in the presence of structured temporal behavior such as
cyclostationarity, where local statistics may vary systematically over time
without indicating any meaningful regime change.

To address this challenge, we perform consistency testing in a learned representation space.
Let $f_\theta : \mathbb{R}^L \rightarrow \mathbb{R}^D$ denote a pretrained time series encoder.
Each window $x_i$ is mapped to an embedding
\vspace{-0.5em}
\begin{equation}
z_i = f_\theta(x_i),
\end{equation}
where the encoder is treated as a fixed feature extractor and is not fine-tuned for the
monitoring task.
The same encoder is applied to both reference and test windows, enabling comparison within a common embedding space. Algorithm~\ref{alg:conformal} summarizes the overall reference-relative consistency testing
procedure.
\vspace{-0.5em}
\subsection{Auxiliary Sensitivity Alignment via Generic Statistical Priors}

The embeddings produced by pretrained time-series encoders reside in a
high-dimensional space $\mathbb{R}^D$ that entangles multiple factors of
variation, including local shape, temporal semantics, and noise\cite{wilinski2025exploring}.
While such richness is desirable for general-purpose tasks, it can be
problematic for reference-relative process control, where the primary interest
lies in detecting deviations from statistical stability.

To address this, we introduce an auxiliary alignment step that identifies
directions in the embedding space that are broadly sensitive to fundamental
statistical changes.
Crucially, this step is not designed to learn specific defects or failure modes.
Instead, it aims to orient the representation space toward a low-dimensional
subspace in which distributional deviations are more easily detectable.

\paragraph{Generic statistical deviation primitives.}
We construct an auxiliary synthetic pool consisting of windowed time series
that instantiate basic \emph{primitives of statistical deviation}, including
changes in mean, variance, trend, and unit-root behavior.
These primitives are not intended to model realistic faults in detail.
Rather, they serve as canonical building blocks, as a wide range of complex
real-world anomalies can ultimately be decomposed into combinations of such
fundamental statistical changes.

Let $z_i = f_\theta(x_i) \in \mathbb{R}^D$ denote the embeddings of these
auxiliary windows under the same pretrained encoder used elsewhere in the paper.
Each window is weakly labeled according to whether it exhibits a stable or
statistically deviated behavior at the level of these primitives.

\paragraph{LDA as a sensitivity filter, not a classifier.}
We use linear discriminant analysis (LDA) to derive a projection
$\Pi : \mathbb{R}^D \rightarrow \mathbb{R}^d$ from the auxiliary pool.
Importantly, LDA is not employed here as a classifier.
Instead, it acts as a linear filter that identifies directions in embedding
space along which distributional shifts induced by basic statistical deviations
are most pronounced.

From this perspective, the projection $\Pi$ can be interpreted as extracting
a \emph{direction of sensitivity} in the representation space, rather than a
decision boundary tied to any specific defect.
The resulting low-dimensional representation emphasizes stability-relevant
variation while suppressing components that are largely invariant under
statistical change.

\paragraph{Generalization and experimental usage.}
The projection $\Pi$ is learned once from the auxiliary synthetic pool and is
then fixed across all experiments.
It does not depend on the reference set, calibration set, or test data of any
particular experiment, nor does it vary across different data-generating
processes.
This design ensures that $\Pi$ cannot overfit to specific regimes or failure
patterns and instead captures a generic axis of stability versus instability.

When auxiliary supervision is unavailable or undesired, the framework naturally
reduces to unsupervised projections (e.g., PCA) or the identity mapping.
We evaluate these alternatives in ablation studies and show that sensitivity
alignment via generic statistical priors substantially improves the efficiency
and robustness of subsequent reference modeling in Appendix \ref{app:proj}.

\subsection{Reference Modeling and Scoring}

We model the reference distribution implicitly based on projected reference embeddings.
Specifically, we treat
$\{\tilde{z}_i : x_i \in \mathcal{R}_{\text{fit}}\}$
as samples from an unknown distribution characterizing normal operation,
which may include both stationary and cyclostationary behavior.

We fit a nonparametric density estimator $\hat{p}(\cdot)$ on these samples.
The consistency score of a window $x$ is defined as the negative log-density
\begin{equation}
s(x) = -\log \hat{p}\!\left( \Pi(f_\theta(x)) \right).
\end{equation}

This score reflects how typical the window is relative to the empirical reference regime.
Scores are computed independently for each window, enabling window-level monitoring without
reliance on temporal context or segmentation.

In all experiments, we model the reference distribution using a Gaussian
kernel density estimator (KDE) in the projected embedding space.
The kernel bandwidth is selected automatically via Silverman’s rule of thumb,
without tuning on test data.
We found this simple choice to be sufficient in low-dimensional projected spaces.
\subsection{Calibration and Decision Thresholds}
\label{sec:calibration}

Given a collection of reference windows representing a context of interest,
our goal is to produce decision-ready statistics whose operating characteristics
are interpretable at a fixed false-alarm rate.
Rather than relying on parametric assumptions,
we calibrate detection scores using a conformal-style procedure based on an
empirical reference distribution\cite{inbook}.
In this framework, calibration determines either a decision threshold
$\tau(\alpha)$ or an equivalent p-value such that the probability of a false alarm
under the null does not exceed a prescribed level $\alpha$.

Let $s(x)$ denote the reference-relative detection score for a window $x$,
where larger values indicate stronger deviation from the reference distribution.
Under global calibration, the conformal p-value for a test window $x$ is defined as
\begin{equation}
p_{\text{global}}(x)
=
\frac{
1 + \sum_{i=1}^{n_{\text{ref}}}
\mathbf{1}\!\left[ s(x_i) \ge s(x) \right]
}{
n_{\text{ref}} + 1
},
\end{equation}
where $\{x_i\}_{i=1}^{n_{\text{ref}}}$ denotes the reference calibration windows.
This definition is standard and enjoys finite-sample validity under the assumption
that calibration and test scores are exchangeable.

In window-based time series analysis, however, detection scores often depend
systematically on the window length.
When reference and test windows of heterogeneous lengths are pooled together during
calibration, the exchangeability assumption can be violated,
leading to length-dependent miscalibration:
some window lengths may become overly liberal, while others become overly conservative,
even when evaluated at the same nominal false-alarm rate.

To address this issue, we adopt a length-conditional calibration strategy.
Let $\mathcal{S}_{\mathrm{cal}}^{(\ell)}$ denote the set of calibration scores
associated with reference windows of length $\ell$,
\[
\mathcal{S}_{\mathrm{cal}}^{(\ell)}
=
\{ s(x_i) \;:\; \mathrm{len}(x_i) = \ell \},
\qquad
n_\ell = \lvert \mathcal{S}_{\mathrm{cal}}^{(\ell)} \rvert .
\]
For a test window $x$ of length $\ell$, the length-conditional conformal p-value is
defined as 
\begin{equation}
p_{\ell}(x)
=
\frac{
1 + \sum_{i:\,\mathrm{len}(x_i)=\ell}
\mathbf{1}\!\left[ s(x_i) \ge s(x) \right]
}{
n_{\ell} + 1
}.
\end{equation}
The corresponding length-conditional decision threshold $\tau_{\ell}(\alpha)$
is obtained by inverting this p-value, and a detection is declared whenever
$p_{\ell}(x) \le \alpha$, equivalently $s(x) \ge \tau_{\ell}(\alpha)$.
Global calibration, obtained by pooling reference windows of all lengths,
is retained as a baseline for comparison.
Length-conditional calibration can be viewed as a refinement of this procedure,
ensuring that the calibration step respects the structural heterogeneity introduced
by varying window lengths.
As shown in Table~\ref{tab:calibration_lengthwise},
global calibration exhibits pronounced length-dependent deviations from the nominal
false-alarm rate, whereas length-conditional calibration restores consistent type-I
error control across all window lengths.
Beyond a single operating point, QQ plots of null p-values further confirm that
length-conditional calibration yields near-uniform p-values over the full
significance range, supporting reliable decision-making at fixed false-alarm rates.

\subsection{Practical Considerations}

The proposed framework naturally supports variable sequence lengths,
as reference and test windows need not share the same length provided the encoder can map
them to fixed-dimensional representations.
Each window is evaluated independently, enabling application to streaming, batch,
or retrospective analysis without modification.

While we focus on a specific instantiation in this work,
the framework is modular.
Alternative encoders, projection operators, density estimators,
or calibration procedures may be substituted without altering the underlying formulation.
Implementation details and hyperparameter settings are provided in the Appendix \ref{app:imp_detail}.

\section{Experiments}
\label{sec:experiments}
\subsection{Experimental Setup and Evaluation Protocol}

We evaluate the proposed framework in a window-level monitoring setting, where the objective is to determine whether individual time series windows are consistent with a given reference regime. All experiments follow a reference-relative protocol: a reference set of windows is first constructed to represent normal operation, and test windows are then evaluated independently against this reference.

\paragraph{Window construction.}
Time series are segmented into local windows, which serve as the basic units of analysis. Unless otherwise specified, windows are extracted without overlap. Throughout the main experiments, we use a \emph{mixed-length} setting by default: window length $L$ varies across samples, while each window is embedded into a fixed-dimensional representation by a pretrained encoder.
To ensure valid calibration under variable lengths, we perform calibration \emph{separately for each length bucket} (i.e., conditional on $L$), and report aggregated performance across lengths. Fixed-length results are provided in the Appendix \ref{app:fixed}.
\vspace{-0.5em}
\paragraph{Reference and test regimes.}
For each experiment, the reference set consists exclusively of windows drawn from a stable operating regime. In our synthetic setting, this regime is defined as a mixture of (i) stationary AR(1) windows and (ii) \emph{stable periodic-template windows}, where within-window dynamics repeat an unknown template with a fixed period-to-window ratio. Concretely, for a window length $L$, we sample $k \sim \mathrm{Unif}\{1,2,4,8\}$ and set $T = L/k$, ensuring that the cycle-to-window ratio is consistent across reference, calibration, and test splits within each length bucket.
Test windows are drawn either from the same regime (in-control) or from distributions exhibiting structured deviations. For AR(1), deviations include shifts in mean/variance, trends, and unit-root behavior. For periodic-template windows, deviations include template changes (pattern change), amplitude changes (step or drift), and period mismatch, as well as their combinations. Importantly, no labels are used during reference modeling or calibration; labels are only used for evaluation.

\paragraph{Decision protocol.}
Each test window is mapped to a representation, projected using a reference-derived projection, and assigned a scalar consistency score relative to the reference distribution. Scores are then calibrated using the empirical distribution of reference scores to produce decision-ready outputs. Unless stated otherwise, we report results at fixed false-alarm rates by thresholding calibrated scores accordingly. Unless stated otherwise, all experiments use Chronos2\cite{chronos2} as the default pretrained
encoder, and other encoders are evaluated only in dedicated ablation studies.

\paragraph{Evaluation metrics.}
Performance is evaluated using detection power at fixed false-alarm rates, reflecting the probability of correctly identifying out-of-regime windows while controlling the rate of false alarms. To assess robustness across operating points, we additionally report performance over a range of false-alarm rates, summarized using power--false-alarm curves and area-under-the-curve (AUC) metrics. These metrics are chosen to reflect the operational nature of the process control setting rather than pointwise classification accuracy.
\vspace{-0.5em}
\paragraph{Baselines.}
We compare the proposed framework against several classical statistical baselines
that operate directly on raw time series windows.
These baselines are included to contextualize performance relative to established
tests, rather than to match the reference-relative formulation exactly.

\emph{Stationarity tests.}
We consider the augmented Dickey--Fuller (ADF), KPSS, and Phillips--Perron (PP)
tests, applied independently to each window\cite{ADF, kpss, pptest}.
These tests assess stationarity against specific parametric alternatives and
produce window-level decisions based on standard test statistics and thresholds.

\emph{Two-stage stationarity--cyclostationarity baselines.}
Classical stationarity tests are limited in their ability to distinguish
non-stationary behavior from cyclostationary structure at the window level.
To account for this limitation, we construct two-stage baselines that combine
a stationarity test (ADF, KPSS, or PP) with the cyclostationarity test of
Dandawate and Giannakis (DG)~\cite{DG_TEST}.
Specifically, a stationarity test is first applied to each window; if the window
is classified as non-stationary, we subsequently apply the DG test to assess
whether the observed deviation is instead consistent with cyclostationary
structure.
If the DG statistic exceeds the corresponding $\chi^2$ threshold, the window is
reclassified as in-regime.
We denote these baselines as \texttt{ADF+DG}, \texttt{KPSS+DG}, and \texttt{PP+DG},
respectively.
A detailed description of the DG test and its implementation in our experiments
is provided in \ref{app:dg_test}.

This two-stage design reflects the complementary scopes of the tests:
ADF, KPSS, and PP distinguish stationarity from general non-stationarity,
while DG specifically targets cyclostationary alternatives.
\vspace{-1.0em}
\paragraph{Fair implementation of DG baselines.}
To avoid providing oracle knowledge to cyclostationarity tests, the
Dandawate--Giannakis (DG) statistic is evaluated using an estimated cyclic
frequency for each window.
When a stage-1 stationarity test flags a window as non-stationary, we estimate
a candidate cyclic frequency by selecting the dominant non-DC peak of the
window’s periodogram.
If the estimated frequency exceeds a minimum resolvable threshold ($2/L$ for
window length $L$), the DG statistic is computed at that frequency and compared
against the corresponding $\chi^2$ critical value.
This procedure is applied uniformly across all DG-based baselines and does not
use ground-truth periodic parameters.

\subsection{Reference-Relative Monitoring under Cyclostationary Regimes}
\label{sec:view2}

Cyclostationary processes provide a canonical example of stable yet non-stationary
operation.
Although their statistics vary periodically over time, such variation is structured
and repeatable, and therefore constitutes a valid notion of normal behavior from a
process control perspective.
This setting poses a fundamental challenge for classical stationarity-based
diagnostics, which treat any deviation from weak stationarity as evidence of
abnormality.

\paragraph{Experimental setup.}
Reference sets are constructed from stationary AR(1) processes, cyclostationary
processes, or mixtures of both.
Test windows are drawn either from the same reference regime or from distributions
exhibiting structured deviations, including mean shifts, variance shifts,
frequency changes, unit-root behavior, and composite disruptions.
All decisions are made independently at the window level, without access to
temporal ordering or change-point information.
\vspace{-0.5em}
\begin{table*}[t]
\centering
\caption{
\textbf{Window-level regime consistency under stationary and cyclostationary reference regimes.}
AUC (mean $\pm$ std over 10 seeds) aggregated across sequence lengths.
Normal windows are defined by the reference regime; higher AUC indicates better reference-relative discrimination.
}
\label{tab:view2_auc_pooled}
\resizebox{\textwidth}{!}{
\begin{tabular}{l|ccccc|cccccc}
\toprule
 & \multicolumn{5}{c|}{\textbf{AR(1) Reference Regime}} 
 & \multicolumn{6}{c}{\textbf{Cyclostationary Reference Regime}} \\
\textbf{Method}
 & Compound & Mean & Trend & Unit-root & Variance
 & Compound & Frequency & Mean & Compound-Structural & Trend & Variance \\
\midrule
\textbf{Proposed}
 & \textbf{0.993} $\pm$ 0.002
 & \textbf{0.992} $\pm$ 0.001
 & \textbf{0.992} $\pm$ 0.002
 & \textbf{0.988} $\pm$ 0.003
 & \textbf{0.988} $\pm$ 0.002
 & \textbf{0.980} $\pm$ 0.003
 & \textbf{0.987} $\pm$ 0.002
 & \textbf{0.974} $\pm$ 0.005
 & \textbf{0.989} $\pm$ 0.002
 & \textbf{0.967} $\pm$ 0.005
 & \textbf{0.982} $\pm$ 0.004 \\
\midrule
ClaSP (CPD)
 & 0.516 $\pm$ 0.018
 & 0.500 $\pm$ 0.011
 & 0.505 $\pm$ 0.017
 & 0.568 $\pm$ 0.013
 & 0.509 $\pm$ 0.018
 & 0.582 $\pm$ 0.016
 & 0.854 $\pm$ 0.015
 & 0.738 $\pm$ 0.025
 & 0.850 $\pm$ 0.014
 & 0.553 $\pm$ 0.015
 & 0.580 $\pm$ 0.018 \\
CUSUM (CPD)
 & 0.339 $\pm$ 0.009
 & 0.392 $\pm$ 0.008
 & 0.301 $\pm$ 0.008
 & 0.612 $\pm$ 0.011
 & 0.277 $\pm$ 0.002
 & 0.700 $\pm$ 0.011
 & 0.709 $\pm$ 0.009
 & 0.682 $\pm$ 0.010
 & 0.731 $\pm$ 0.005
 & 0.686 $\pm$ 0.013
 & 0.700 $\pm$ 0.014 \\
\midrule
ADF + DG
 & 0.691 $\pm$ 0.023
 & 0.937 $\pm$ 0.004
 & 0.920 $\pm$ 0.008
 & 0.831 $\pm$ 0.013
 & 0.459 $\pm$ 0.007
 & 0.531 $\pm$ 0.011
 & 0.510 $\pm$ 0.012
 & 0.532 $\pm$ 0.015
 & 0.512 $\pm$ 0.016
 & 0.534 $\pm$ 0.011
 & 0.531 $\pm$ 0.012 \\
KPSS + DG
 & 0.746 $\pm$ 0.017
 & 0.944 $\pm$ 0.003
 & 0.941 $\pm$ 0.005
 & 0.807 $\pm$ 0.009
 & 0.478 $\pm$ 0.008
 & 0.524 $\pm$ 0.013
 & 0.498 $\pm$ 0.013
 & 0.502 $\pm$ 0.009
 & 0.498 $\pm$ 0.011
 & 0.528 $\pm$ 0.007
 & 0.520 $\pm$ 0.010 \\
PP + DG
 & 0.500 $\pm$ 0.010
 & 0.781 $\pm$ 0.017
 & 0.657 $\pm$ 0.008
 & 0.823 $\pm$ 0.013
 & 0.436 $\pm$ 0.003
 & 0.531 $\pm$ 0.013
 & 0.531 $\pm$ 0.022
 & 0.549 $\pm$ 0.013
 & 0.531 $\pm$ 0.012
 & 0.547 $\pm$ 0.011
 & 0.535 $\pm$ 0.015 \\
\bottomrule
\end{tabular}
}
\vspace{-1.0em}
\end{table*}

\paragraph{Overall discrimination performance.}
\cref{tab:view2_auc_pooled} summarizes pooled ROC-AUC results across deviation types.
The proposed method achieves consistently high discrimination performance across
both stationary and cyclostationary reference regimes, remaining close to perfect
in most settings.
Notably, strong performance is maintained for frequency-based and composite
deviations that preserve local stability yet violate the reference regime structure.
\vspace{-0.5em}
\paragraph{Comparison to change-point detection baselines.}
Change-point detection (CPD) methods such as CUSUM and ClaSP\cite{Hawkins1998CumulativeSC,clasp2023}
perform poorly in the window-level regime consistency setting, often approaching random guessing.
To enable a fair comparison, we adapt CPD methods to produce window-level scores by applying each
detector independently to individual windows and using the resulting test statistic
(or maximum detector response within the window) as a proxy for deviation strength.

Despite this adaptation, CPD methods remain ineffective.
This behavior is expected, as CPD algorithms are fundamentally designed to identify
temporal boundaries between regimes by exploiting sequential context,
rather than to assess distributional consistency of isolated windows relative to a reference set.
When applied to standalone windows without access to pre- and post-change context,
their core modeling assumptions are violated, leading to unstable or uninformative scores.
\vspace{-0.5em}
\paragraph{Implications.}
These results demonstrate that the proposed framework decouples the notion of
normality from fixed parametric assumptions.
By defining normal operation empirically through a reference set and performing
consistency testing in a learned representation space, the method naturally
accommodates cyclostationary behavior as a stable regime.
\vspace{-0.5em}
\subsection{Additional Results under Stationary Reference Regimes}
\label{sec:view1}

While cyclostationary reference regimes expose fundamental limitations of
classical stationarity diagnostics, it is equally important to verify that the
proposed framework does not incur unnecessary loss in settings that are
well-aligned with traditional stationarity assumptions.
We therefore include a complementary evaluation under stationary AR(1)
reference regimes.
\vspace{-0.5em}

\paragraph{Experimental setting.}
In this experiment, reference windows are drawn from stationary AR(1) processes,
and test windows consist of either stationary AR(1) windows or windows exhibiting
classical non-stationary deviations such as mean shifts, variance changes, trends,
or unit-root behavior.
All methods operate at the window level.
For the proposed approach, decisions are obtained via conformal calibration at
fixed false-alarm rates $\alpha \in \{0.01, 0.05\}$.
Results are aggregated across all sequence lengths and reported as mean $\pm$ std
over 20 random seeds.
\vspace{-0.5em}

\paragraph{Results and discussion.}
Table~\ref{tab:view1_op_avg_method} reports the true positive rate (TPR) and false
positive rate (FPR) at the two operating points.
The proposed method achieves near-perfect detection power at both $\alpha=0.01$
and $\alpha=0.05$, while maintaining controlled false-alarm rates.
In particular, at $\alpha=0.01$, the method attains a TPR of $0.984$ with an FPR
well below the nominal level, indicating conservative but reliable behavior.

Classical stationarity tests exhibit the expected trade-offs.
ADF and KPSS achieve moderate detection power but with elevated or unstable false
alarm rates depending on the operating point, while PP is highly conservative,
yielding low FPR at the expense of substantially reduced detection power.
These results reflect the well-known sensitivity of classical tests to specific
assumptions and window lengths.

Overall, this experiment confirms that the reference-relative formulation strictly
generalizes classical stationarity diagnostics.
When the reference regime satisfies traditional stationarity assumptions, the
proposed method retains strong performance and calibrated operating characteristics,
while remaining applicable to broader regimes considered elsewhere in the paper.
\vspace{-0.5em}
\begin{table}[t]
\centering
\caption{
\textbf{Operating-point performance under stationary AR(1) reference regimes.}
Window-level detection results at fixed false-alarm rates ($\alpha \in \{0.01, 0.05\}$),
aggregated across all sequence lengths.
True positive rate (TPR) and false positive rate (FPR) are reported as mean $\pm$ std
over 20 random seeds.
}
\label{tab:view1_op_avg_method}
\resizebox{\columnwidth}{!}{
\begin{tabular}{l|cc|cc}
\toprule
& \multicolumn{2}{c|}{$\alpha=0.01$} & \multicolumn{2}{c}{$\alpha=0.05$} \\
\textbf{Method} & \textbf{TPR} $\uparrow$ & \textbf{FPR} $\downarrow$ & \textbf{TPR} $\uparrow$ & \textbf{FPR} $\downarrow$ \\
\midrule
\textbf{Proposed} & \textbf{0.984} $\pm$ 0.005 & \textbf{0.006} $\pm$ 0.003 & \textbf{0.998} $\pm$ 0.002 & 0.063 $\pm$ 0.012 \\
ADF & 0.744 $\pm$ 0.008 & 0.045 $\pm$ 0.006 & 0.699 $\pm$ 0.009 & 0.019 $\pm$ 0.005 \\
KPSS & 0.669 $\pm$ 0.009 & 0.018 $\pm$ 0.004 & 0.729 $\pm$ 0.010 & 0.091 $\pm$ 0.009 \\
PP & 0.526 $\pm$ 0.012 & 0.029 $\pm$ 0.005 & 0.456 $\pm$ 0.011 & \textbf{0.005} $\pm$ 0.002 \\
\bottomrule
\end{tabular}
\vspace{-1.0em}
}
\end{table}
\begin{figure*}[t]
\centering
\begin{subfigure}[t]{0.32\textwidth}
    \centering
    \includegraphics[width=\linewidth]{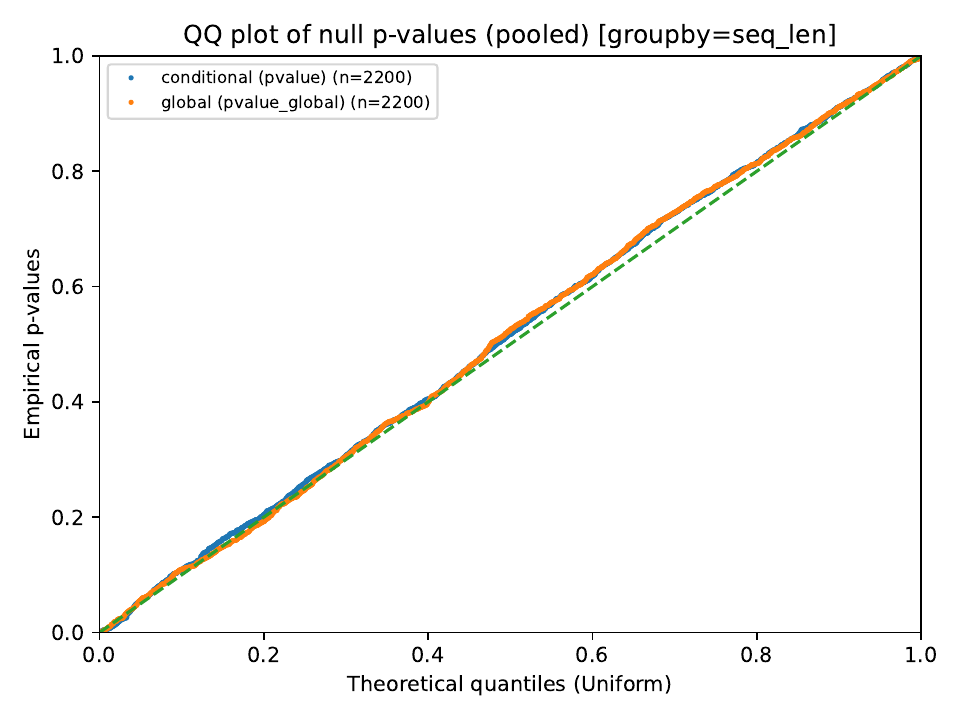}
    \caption{Globally pooled calibration}
    \label{fig:qq_all}
\end{subfigure}
\hfill
\begin{subfigure}[t]{0.32\textwidth}
    \centering
    \includegraphics[width=\linewidth]{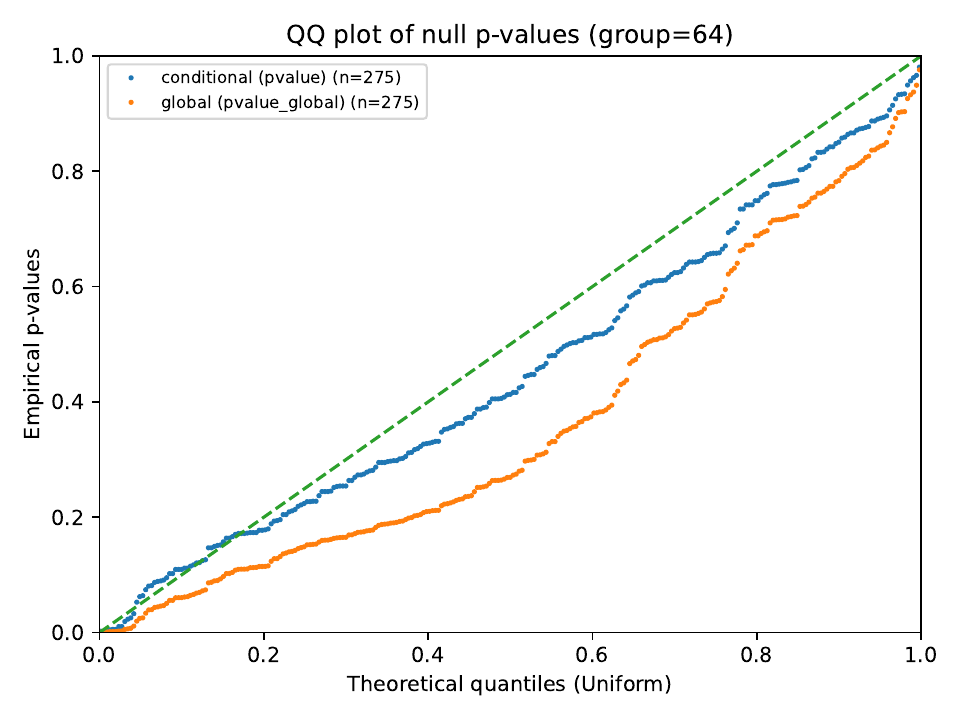}
    \caption{Length-conditional ($\texttt{seq\_len}=64$)}
    \label{fig:qq_64}
\end{subfigure}
\hfill
\begin{subfigure}[t]{0.32\textwidth}
    \centering
    \includegraphics[width=\linewidth]{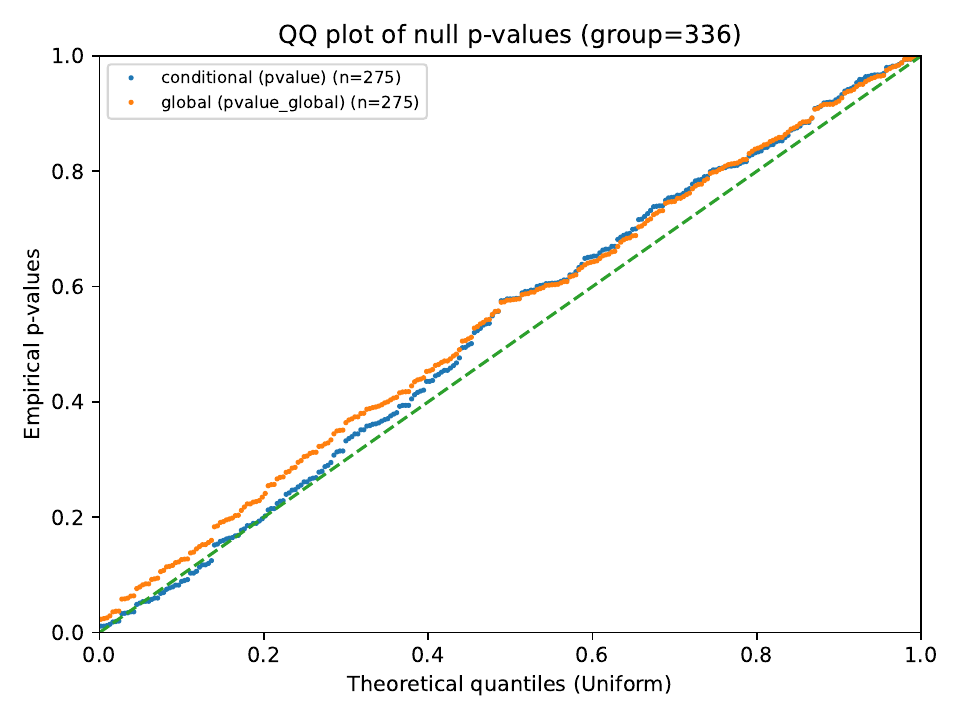}
    \caption{Length-conditional ($\texttt{seq\_len}=336$)}
    \label{fig:qq_336}
\end{subfigure}

\vspace{-0.3em}
\caption{
QQ plots of null p-values under mixed window lengths.
\textbf{Left:} Globally pooled calibration appears approximately uniform marginally,
but masks systematic length-dependent miscalibration.
\textbf{Middle and right:} Length-conditional calibration yields near-uniform p-values
for both short and long windows, explaining the stable type-I error control across lengths
reported in Table~\ref{tab:calibration_lengthwise}.
}
\label{fig:qqplots}
\vspace{-0.5em}
\end{figure*}

\begin{table}[t]
\centering
\caption{
Length-wise calibration diagnostics under mixed window lengths.
We report the empirical type-I error at $\alpha=0.05$ and the maximum deviation
from nominal calibration,
$\sup_{\alpha}\lvert \widehat{\mathbb{P}}(p \le \alpha) - \alpha \rvert$,
for global pooling and length-conditional calibration.
}
\label{tab:calibration_lengthwise}
\resizebox{0.95\columnwidth}{!}{
\begin{tabular}{c|cc|cc}
\toprule
\texttt{seq\_len}
& \multicolumn{2}{c|}{Type-I error @ $\alpha=0.05$}
& \multicolumn{2}{c}{Max. calibration deviation} \\
& Global & Conditional
& Global & Conditional \\
\midrule
64  & 0.0800 & 0.0436 & 0.1818 & 0.0256 \\
96  & 0.0509 & 0.0400 & 0.0800 & 0.0339 \\
128 & 0.0400 & 0.0400 & 0.0366 & 0.0458 \\
192 & 0.0436 & 0.0400 & 0.0327 & 0.0220 \\
256 & 0.0545 & 0.0582 & 0.0691 & 0.0406 \\
336 & 0.0255 & 0.0473 & 0.0454 & 0.0162 \\
512 & 0.0364 & 0.0473 & 0.0291 & 0.0218 \\
720 & 0.0400 & 0.0436 & 0.0442 & 0.0220 \\
\bottomrule
\end{tabular}
}
\end{table}

\subsection{Ablation Study}
\label{sec:ablation}
We conclude with an ablation study examining the robustness of the proposed framework to key implementation choices.
\vspace{-0.5em}
\begin{table}[t]
\centering
\caption{
Ablation study on KDE bandwidth selection.
We report performance under different bandwidth rules and scaling factors.
Results show stable performance across a wide range of bandwidth choices.
}
\label{tab:ablation_bw}
\resizebox{0.95\columnwidth}{!}{
\begin{tabular}{l c c c c c c}
\toprule
Rule of Thumbs & Mult. & ROC-AUC & PR-AUC & $\tau_{\alpha=0.05}$ & F1 ($\alpha=0.05$) & Recall \\
\midrule
\multirow{4}{*}{\large Scott}     & 0.50 & 0.9514 & 0.9670 & 2.5267 & 0.9241 & 0.8882 \\
                                  & 1.00 & 0.9519 & 0.9672 & 2.4551 & 0.9243 & 0.8905 \\
                                  & 1.50 & 0.9525 & 0.9676 & 2.3665 & 0.9244 & 0.8923 \\
                                  & 2.00 & 0.9524 & 0.9679 & 2.2672 & 0.9243 & 0.8936 \\
\midrule
\multirow{4}{*}{\large Silverman} & 0.50 & 0.9513 & 0.9670 & 2.5371 & 0.9240 & 0.8873 \\
                                  & 1.00 & 0.9518 & 0.9671 & 2.4725 & 0.9243 & 0.8905 \\
                                  & 1.50 & 0.9523 & 0.9675 & 2.3869 & 0.9244 & 0.8918 \\
                                  & 2.00 & 0.9526 & 0.9678 & 2.3094 & 0.9245 & 0.8932 \\
\bottomrule
\end{tabular}
}
\end{table}
\paragraph{Effect of KDE bandwidth.}
\cref{tab:ablation_bw} shows performance under different bandwidth selection rules and scaling factors.
Across both Silverman's and Scott's rules\cite{silverman, scott}, and over a wide range of bandwidth multipliers, performance remains highly stable.
While the calibrated decision threshold varies smoothly with the degree of smoothing, metrics such as ROC-AUC, PR-AUC, and F1-score remain nearly unchanged.
This indicates that the method does not rely on fine-grained bandwidth tuning.
\vspace{-1.0em}
\begin{table}[t]
\centering
\caption{
Ablation study on the size of the reference set ($n_{\text{ref}}$).
All metrics are reported at a fixed false-alarm rate $\alpha=0.05$ using
length-conditional calibration.
Performance remains stable across a wide range of reference sizes, with
both detection accuracy and decision thresholds exhibiting minimal sensitivity
to $n_{\text{ref}}$.
}
\label{tab:ablation_nref}
\resizebox{0.95\columnwidth}{!}{
\begin{tabular}{l|cccccc}
\toprule
$n_{\text{ref}}$
& ROC-AUC
& PR-AUC
& $\tau$ $(\alpha=0.05)$
& F1-Score
& Precision
& Recall \\
\midrule
5{,}000
 & 0.9816 & 0.9788 & 3.3722 & 0.9353 & 0.9550 & 0.9164 \\
10{,}000
 & 0.9810 & 0.9745 & 3.1031 & 0.9315 & 0.9551 & 0.9091 \\
15{,}000
 & \textbf{0.9880} & 0.9820 & 3.1457 & \textbf{0.9475} & 0.9560 & \textbf{0.9391} \\
20{,}000
 & 0.9859 & 0.9827 & 3.1346 & 0.9399 & 0.9486 & 0.9314 \\
25{,}000
 & 0.9827 & 0.9799 & 3.1794 & 0.9253 & 0.9541 & 0.8982 \\
50{,}000
 & 0.9827 & 0.9791 & 3.1555 & 0.9286 & 0.9505 & 0.9077 \\
75{,}000
 & 0.9854 & 0.9798 & 3.1611 & 0.9386 & \textbf{0.9566} & 0.9214 \\
100{,}000
 & 0.9865 & \textbf{0.9833} & 3.1357 & 0.9420 & 0.9547 & 0.9295 \\
\bottomrule
\end{tabular}
}
\end{table}
\paragraph{Effect of reference set size.}
\cref{tab:ablation_nref} examines the impact of the number of reference windows.
As expected, performance improves with larger reference sets, but the method
remains effective even with relatively small reference sizes.
Importantly, conformal calibration continues to enforce false-alarm control,
and performance degrades gracefully rather than collapsing in low-sample regimes.
\vspace{-0.5em}
\paragraph{Summary.}
Overall, the ablation results demonstrate that the proposed reference-relative
framework is robust to key design choices, including KDE bandwidth selection
and reference set size.
This robustness is essential for practical process control settings, where
hyperparameter tuning opportunities may be limited and operating conditions may
vary across deployments.
\vspace{-1.0em}
\section{Conclusion}

In this work, we revisited window-level time series monitoring from a process control perspective, arguing that practical monitoring is inherently \emph{reference-relative}. Instead of testing predefined parametric hypotheses such as stationarity, we proposed a framework that assesses whether individual windows remain distributionally consistent with an empirical reference regime. By operating in a learned representation space, modeling the reference distribution nonparametrically, and applying conformal calibration, the proposed method provides decision-ready, finite-sample-valid consistency assessments without relying on deviation-specific assumptions.

Our experiments demonstrate that this formulation naturally accommodates both stationary and cyclostationary regimes as valid notions of normal operation. We showed that classical stationarity-based diagnostics and change-point detection methods struggle in settings where the reference regime itself exhibits structured non-stationarity. In contrast, the proposed framework achieves strong, well-calibrated performance across a wide range of deviations while remaining robust to implementation choices. These results suggest that reference-relative diagnostics provide a more rigorous and flexible foundation for process control in modern time series systems.
\vspace{-0.8em}
\paragraph{Limitations.}
Our framework entails several limitations that outline directions for future research.
First, as a reference-relative approach, its effectiveness is inherently bounded by the
representativeness of the reference set $\mathcal{R}$.
Since classical SPC relies on establishing control limits from historical
reference distributions~\cite{montgomery2020introduction}, any mismatch where the reference windows
fail to capture the full support of the normal operating regime---for instance,
missing rare but benign cyclostationary phases---may yield elevated false positive rates.
This highlights the practical necessity of careful reference construction.

Second, while we leverage the zero-shot capabilities of pretrained encoders
like Chronos2 and MOMENT, the assumption of ``universal'' representations may
falter under extreme domain shifts where the encoder's pretraining corpus
diverges significantly from the monitored signal~\cite{chronos2, moment}.
Although our ablation studies demonstrate robustness, performance is ultimately
contingent on the quality of the frozen embedding space.

Finally, our experimental validation primarily relies on controlled synthetic
regimes (AR processes and periodic templates) to rigorously verify statistical
calibration and Type-I error control.
While this isolates the framework's theoretical properties, extending evaluation
to large-scale, noisy, and unlabeled real-world benchmarks remains an essential
step to confirm practical utility in diverse deployment scenarios.

\newpage
\section*{Impact Statement}

This work studies window-level consistency diagnostics for time series monitoring,
with a focus on reference-relative decision making and calibrated false-alarm control.
The proposed framework is intended for use in benign monitoring and quality-control
settings, such as industrial process supervision, system health monitoring,
and exploratory analysis of temporal data.

The methodology does not involve human subjects, personal data, or automated decision
making affecting individuals.
As such, we do not anticipate direct negative societal impacts arising from this work.
Potential misuse could occur if the proposed diagnostics were applied without proper
domain knowledge or reference construction, leading to misleading conclusions about
system behavior.
However, this risk is common to statistical monitoring tools in general and can be
mitigated through careful reference selection and validation.

We believe the primary impact of this work is methodological.
By reframing classical stationarity and cyclostationarity diagnostics as
reference-relative consistency testing in learned representation spaces,
this work may help bridge statistical process control and modern time-series
representation learning.
More broadly, the emphasis on calibrated, reference-dependent decisions may encourage
the responsible deployment of monitoring systems with explicit control over false
alarms, rather than reliance on ad hoc anomaly scores.

\bibliographystyle{icml2025}
\bibliography{works}

@String(AAAI = {AAAI})

@inproceedings{moment,
  title={MOMENT: A Family of Open Time-series Foundation Models},
  author={Mononito Goswami and Konrad Szafer and Arjun Choudhry and Yifu Cai and Shuo Li and Artur Dubrawski},
  booktitle={International Conference on Machine Learning},
  year={2024}
}

@article{moirai_moe,
  title={Moirai-MoE: Empowering Time Series Foundation Models with Sparse Mixture of Experts},
  author={Liu, Xu and Liu, Juncheng and Woo, Gerald and Aksu, Taha and Liang, Yuxuan and Zimmermann, Roger and Liu, Chenghao and Savarese, Silvio and Xiong, Caiming and Sahoo, Doyen},
  journal={arXiv preprint arXiv:2410.10469},
  year={2024}
}

@inproceedings{moirai,
  title={Unified Training of Universal Time Series Forecasting Transformers},
  author={Woo, Gerald and Liu, Chenghao and Kumar, Akshat and Xiong, Caiming and Savarese, Silvio and Sahoo, Doyen},
  booktitle={Forty-first International Conference on Machine Learning},
  year={2024}
}

@ARTICLE{DG_TEST,
  author={Dandawate, A.V. and Giannakis, G.B.},
  journal={IEEE Transactions on Signal Processing}, 
  title={Statistical tests for presence of cyclostationarity}, 
  year={1994},
  volume={42},
  number={9},
  pages={2355-2369},
  doi={10.1109/78.317857}}

@article{chronos1,
  title={Chronos: Learning the Language of Time Series},
  author={Ansari, Abdul Fatir and Stella, Lorenzo and Turkmen, Caner and Zhang, Xiyuan and Mercado, Pedro and Shen, Huibin and Shchur, Oleksandr and Rangapuram, Syama Syndar and Pineda Arango, Sebastian and Kapoor, Shubham and Zschiegner, Jasper and Maddix, Danielle C. and Mahoney, Michael W. and Torkkola, Kari and Gordon Wilson, Andrew and Bohlke-Schneider, Michael and Wang, Yuyang},
  journal={Transactions on Machine Learning Research},
  issn={2835-8856},
  year={2024},
  url={https://openreview.net/forum?id=gerNCVqqtR}
}

@article{chronos2,
  title        = {Chronos-2: From Univariate to Universal Forecasting},
  author       = {Abdul Fatir Ansari and Oleksandr Shchur and Jaris Küken and Andreas Auer and Boran Han and Pedro Mercado and Syama Sundar Rangapuram and Huibin Shen and Lorenzo Stella and Xiyuan Zhang and Mononito Goswami and Shubham Kapoor and Danielle C. Maddix and Pablo Guerron and Tony Hu and Junming Yin and Nick Erickson and Prateek Mutalik Desai and Hao Wang and Huzefa Rangwala and George Karypis and Yuyang Wang and Michael Bohlke-Schneider},
  journal      = {arXiv preprint arXiv:2510.15821},
  year         = {2025},
  url          = {https://arxiv.org/abs/2510.15821}
}

@article{ADF,
 ISSN = {01621459, 1537274X},
 URL = {http://www.jstor.org/stable/2286348},
 author = {David A. Dickey and Wayne A. Fuller},
 journal = {Journal of the American Statistical Association},
 number = {366},
 pages = {427--431},
 publisher = {[American Statistical Association, Taylor & Francis, Ltd.]},
 title = {Distribution of the Estimators for Autoregressive Time Series With a Unit Root},
 urldate = {2025-12-29},
 volume = {74},
 year = {1979}
}

@article{kpss,
title = "Testing the null hypothesis of stationarity against the alternative of a unit root. How sure are we that economic time series have a unit root?",
author = "Denis Kwiatkowski and Phillips, \{Peter C.B.\} and Peter Schmidt and Yongcheol Shin",
year = "1992",
doi = "10.1016/0304-4076(92)90104-Y",
language = "English",
volume = "54",
pages = "159--178",
journal = "Journal of Econometrics",
issn = "0304-4076",
number = "1-3",
}

@article{Gardner_1986, title={The spectral correlation theory of cyclostationary time-series}, volume={11}, ISSN={0165-1684}, DOI={10.1016/0165-1684(86)90092-7}, number={1}, journal={Signal Processing}, author={Gardner, William A.}, year={1986}, month=july, pages={13–36} }

@article{Napolitano_2016, title={Cyclostationarity: New trends and applications}, volume={120}, ISSN={0165-1684}, DOI={10.1016/j.sigpro.2015.09.011}, abstractNote={A concise survey of the literature on cyclostationarity of the last 10 years is presented and an extensive bibliography included. The problems of statistical function estimation, signal detection, and cycle frequency estimation are reviewed. Applications in communications are addressed. In particular, spectrum sensing and signal classification for cognitive radio, source location, MMSE filtering, and compressive sensing are discussed. Limits to the applicability of the cyclostationary signal processing and generalizations of cyclostationarity to overcome these limits are addressed in the companion paper “Cyclostationarity: Limits and generalizations”.}, journal={Signal Processing}, author={Napolitano, Antonio}, year={2016}, month=mar, pages={385–408} }

@article{TiRex, title={TiRex: Zero-Shot Forecasting Across Long and Short Horizons with Enhanced In-Context Learning}, url={http://arxiv.org/abs/2505.23719}, DOI={10.48550/arXiv.2505.23719}, abstractNote={In-context learning, the ability of large language models to perform tasks using only examples provided in the prompt, has recently been adapted for time series forecasting. This paradigm enables zero-shot prediction, where past values serve as context for forecasting future values, making powerful forecasting tools accessible to non-experts and increasing the performance when training data are scarce. Most existing zero-shot forecasting approaches rely on transformer architectures, which, despite their success in language, often fall short of expectations in time series forecasting, where recurrent models like LSTMs frequently have the edge. Conversely, while LSTMs are well-suited for time series modeling due to their state-tracking capabilities, they lack strong in-context learning abilities. We introduce TiRex that closes this gap by leveraging xLSTM, an enhanced LSTM with competitive in-context learning skills. Unlike transformers, state-space models, or parallelizable RNNs such as RWKV, TiRex retains state-tracking, a critical property for long-horizon forecasting. To further facilitate its state-tracking ability, we propose a training-time masking strategy called CPM. TiRex sets a new state of the art in zero-shot time series forecasting on the HuggingFace benchmarks GiftEval and Chronos-ZS, outperforming significantly larger models including TabPFN-TS (Prior Labs), Chronos Bolt (Amazon), TimesFM (Google), and Moirai (Salesforce) across both short- and long-term forecasts.}, note={arXiv:2505.23719 [cs]}, number={arXiv:2505.23719}, publisher={arXiv}, author={Auer, Andreas and Podest, Patrick and Klotz, Daniel and Böck, Sebastian and Klambauer, Günter and Hochreiter, Sepp}, year={2025}, month=may }

@article{
talukder2024totem,
title={{TOTEM}: {TO}kenized Time Series {EM}beddings for General Time Series Analysis},
author={Sabera J Talukder and Yisong Yue and Georgia Gkioxari},
journal={Transactions on Machine Learning Research},
issn={2835-8856},
year={2024},
url={https://openreview.net/forum?id=QlTLkH6xRC},
note={}
}

@book{Cryer_Chan_2008, series={Springer Texts in Statistics}, title={Time Series Analysis: With Applications in R}, ISBN={978-0-387-75959-3}, url={https://books.google.com/books?id=bHke2k-QYP4C}, publisher={Springer New York}, author={Cryer, J.D. and Chan, K.S.}, year={2008}, collection={Springer Texts in Statistics} }

@article{clasp2023,
  title={ClaSP: parameter-free time series segmentation},
  author={Arik Ermshaus and Patrick Sch{\"a}fer and Ulf Leser},
  journal={Data Mining and Knowledge Discovery},
  year={2023},
}

@misc{sjosten2022comparative,
  title={A Comparative Study of the KPSS and ADF Tests in terms of Size and Power},
  author={Sj{\"o}sten, Lina},
  year={2022}
}

@inproceedings{
TNC,
title={Unsupervised Representation Learning for Time Series with Temporal Neighborhood Coding},
author={Sana Tonekaboni and Danny Eytan and Anna Goldenberg},
booktitle={International Conference on Learning Representations},
year={2021},
url={https://openreview.net/forum?id=8qDwejCuCN}
}

@book{montgomery2020introduction,
  title={Introduction to statistical quality control},
  author={Montgomery, Douglas C},
  year={2020},
  publisher={John wiley \& sons}
}

@article{Page1954CONTINUOUSIS,
  title={CONTINUOUS INSPECTION SCHEMES},
  author={E. S. Page},
  journal={Biometrika},
  year={1954},
  volume={41},
  pages={100-115},
  url={https://api.semanticscholar.org/CorpusID:121530032}
}

@article{Roberts2000ControlCT,
  title={Control Chart Tests Based on Geometric Moving Averages},
  author={S. W. Roberts},
  journal={Technometrics},
  year={2000},
  volume={42},
  pages={101 - 97},
  url={https://api.semanticscholar.org/CorpusID:24647878}
}

@article{pptest,
 ISSN = {00063444},
 URL = {http://www.jstor.org/stable/2336182},
 author = {Peter C. B. Phillips and Pierre Perron},
 journal = {Biometrika},
 number = {2},
 pages = {335--346},
 publisher = {[Oxford University Press, Biometrika Trust]},
 title = {Testing for a Unit Root in Time Series Regression},
 urldate = {2026-01-26},
 volume = {75},
 year = {1988}
}

@article{Aminikhanghahi2016ASO,
  title={A survey of methods for time series change point detection},
  author={Samaneh Aminikhanghahi and Diane Joyce Cook},
  journal={Knowledge and Information Systems},
  year={2016},
  volume={51},
  pages={339 - 367},
  url={https://api.semanticscholar.org/CorpusID:15595198}
}

@book{basseville,
author = {Basseville, Michèle and Nikiforov, Igor},
year = {1993},
month = {04},
pages = {},
title = {Detection of Abrupt Change Theory and Application},
volume = {15},
isbn = {0-13-126780-9}
}

@article{truong,
title = {Selective review of offline change point detection methods},
journal = {Signal Processing},
volume = {167},
pages = {107299},
year = {2020},
issn = {0165-1684},
doi = {https://doi.org/10.1016/j.sigpro.2019.107299},
url = {https://www.sciencedirect.com/science/article/pii/S0165168419303494},
author = {Charles Truong and Laurent Oudre and Nicolas Vayatis}
}

@article{chandola,
author = {Chandola, Varun and Banerjee, Arindam and Kumar, Vipin},
title = {Anomaly detection: A survey},
year = {2009},
issue_date = {July 2009},
publisher = {Association for Computing Machinery},
address = {New York, NY, USA},
volume = {41},
number = {3},
issn = {0360-0300},
url = {https://doi.org/10.1145/1541880.1541882},
doi = {10.1145/1541880.1541882},
journal = {ACM Comput. Surv.},
month = jul,
articleno = {15},
numpages = {58}
}

@inproceedings{hundman2018detecting,
author = {Hundman, Kyle and Constantinou, Valentino and Laporte, Christopher and Colwell, Ian and Soderstrom, Tom},
title = {Detecting Spacecraft Anomalies Using LSTMs and Nonparametric Dynamic Thresholding},
year = {2018},
isbn = {9781450355520},
publisher = {Association for Computing Machinery},
address = {New York, NY, USA},
url = {https://doi.org/10.1145/3219819.3219845},
doi = {10.1145/3219819.3219845},
booktitle = {Proceedings of the 24th ACM SIGKDD International Conference on Knowledge Discovery \& Data Mining},
pages = {387–395},
numpages = {9},
location = {London, United Kingdom},
series = {KDD '18}
}

@misc{burger2025distributionfreeprocessmonitoringconformal,
      title={Distribution-Free Process Monitoring with Conformal Prediction}, 
      author={Christopher Burger},
      year={2025},
      eprint={2512.23602},
      archivePrefix={arXiv},
      primaryClass={cs.LG},
      url={https://arxiv.org/abs/2512.23602}, 
}

@article{quionero,
author = {Quionero-Candela, Joaquin and Sugiyama, Masashi and Schwaighofer, Anton and Lawrence, Neil},
year = {2009},
month = {01},
pages = {},
title = {Dataset Shift in Machine Learning},
isbn = {0262170051, 9780262170055}
}

@book{silverman,
  title={Density Estimation for Statistics and Data Analysis},
  author={Silverman, B.W.},
  isbn={9780412246203},
  lccn={86021347},
  series={Chapman \& Hall/CRC Monographs on Statistics \& Applied Probability},
  url={https://books.google.com/books?id=e-xsrjsL7WkC},
  year={1986},
  publisher={Taylor \& Francis}
}

@article{scott,
    author = {Scott, DAVID W.},
    title = {On optimal and data-based histograms},
    journal = {Biometrika},
    volume = {66},
    number = {3},
    pages = {605-610},
    year = {1979},
    month = {12},
    issn = {0006-3444},
    doi = {10.1093/biomet/66.3.605},
    url = {https://doi.org/10.1093/biomet/66.3.605},
    eprint = {https://academic.oup.com/biomet/article-pdf/66/3/605/632347/66-3-605.pdf},
}

@article{Fretheim748,
	author = {Fretheim, Atle and Tomic, Oliver},
	doi = {10.1136/bmjqs-2014-003756},
	eprint = {https://qualitysafety.bmj.com/content/24/12/748.full.pdf},
	issn = {2044-5415},
	journal = {BMJ Quality \& Safety},
	number = {12},
	pages = {748--752},
	publisher = {BMJ Publishing Group Ltd},
	title = {Statistical process control and interrupted time series: a golden opportunity for impact evaluation in quality improvement},
	url = {https://qualitysafety.bmj.com/content/24/12/748},
	volume = {24},
	year = {2015}}

@inproceedings{Hawkins1998CumulativeSC,
  title={Cumulative Sum Charts and Charting for Quality Improvement},
  author={Douglas M. Hawkins and David H. Olwell},
  booktitle={Statistics for Engineering and Physical Science},
  year={1998},
  url={https://api.semanticscholar.org/CorpusID:41296339}
}

@article{math11163444,
	article-number = {3444},
	author = {Benkov{\'a}, Marta and Bedn{\'a}rov{\'a}, Dagmar and Bogdanovsk{\'a}, Gabriela and Pavl{\'\i}{\v c}kov{\'a}, Marcela},
	doi = {10.3390/math11163444},
	issn = {2227-7390},
	journal = {Mathematics},
	number = {16},
	title = {Use of Statistical Process Control for Coking Time Monitoring},
	url = {https://www.mdpi.com/2227-7390/11/16/3444},
	volume = {11},
	year = {2023}}

@inbook{inbook,
author = {Vovk, Vladimir and Gammerman, Alex and Shafer, Glenn},
year = {2005},
month = {01},
pages = {},
title = {Algorithmic Learning in a Random World},
journal = {Algorithmic Learning in a Random World},
doi = {10.1007/b106715}
}

@inproceedings{wilinski2025exploring,
  title={Exploring Representations and Interventions in Time Series Foundation Models},
  author={Micha{\l} Wili{\'n}ski and Mononito Goswami and Willa Potosnak and Nina {\.{Z}}ukowska and Artur Dubrawski},
  booktitle={Forty-second International Conference on Machine Learning},
  year={2025},
  url={https://openreview.net/forum?id=goVzfYtj58}
}

@inproceedings{
li2025online,
title={Online Time Series Forecasting with Theoretical Guarantees},
author={Zijian Li and Changze Zhou and Minghao Fu and Sanjay Manjunath and Fan Feng and Guangyi Chen and Yingyao Hu and Ruichu Cai and Kun Zhang},
booktitle={The Thirty-ninth Annual Conference on Neural Information Processing Systems},
year={2025},
url={https://openreview.net/forum?id=0XCZWAo7wN}
}

@article{gama,
author = {Gama, Jo\~{a}o and \v{Z}liobaitundefined, Indrundefined and Bifet, Albert and Pechenizkiy, Mykola and Bouchachia, Abdelhamid},
title = {A survey on concept drift adaptation},
year = {2014},
issue_date = {April 2014},
publisher = {Association for Computing Machinery},
address = {New York, NY, USA},
volume = {46},
number = {4},
issn = {0360-0300},
url = {https://doi.org/10.1145/2523813},
doi = {10.1145/2523813},
journal = {ACM Comput. Surv.},
month = mar,
articleno = {44},
numpages = {37}
}

@article{online,
	abstractnote = {&lt;p&gt; Autoregressive integrated moving average (ARIMA) is one of the most popular linear models for time series forecasting due to its nice statistical properties and great flexibility. However, its parameters are estimated in a batch manner and its noise terms are often assumed to be strictly bounded, which restricts its applications and makes it inefficient for handling large-scale real data. In this paper, we propose online learning algorithms for estimating ARIMA models under relaxed assumptions on the noise terms, which is suitable to a wider range of applications and enjoys high computational efficiency. The idea of our ARIMA method is to reformulate the ARIMA model into a task of full information online optimization (without random noise terms). As a consequence, we can online estimation of the parameters in an efficient and scalable way. Furthermore, we analyze regret bounds of the proposed algorithms, which guarantee that our online ARIMA model is provably as good as the best ARIMA model in hindsight. Finally, our encouraging experimental results further validate the effectiveness and robustness of our method. &lt;/p&gt;},
	author = {Liu, Chenghao and Hoi, Steven C.H. and Zhao, Peilin and Sun, Jianling},
	doi = {10.1609/aaai.v30i1.10257},
	journal = {Proceedings of the AAAI Conference on Artificial Intelligence},
	month = {Feb.},
	number = {1},
	title = {Online ARIMA Algorithms for Time Series Prediction},
	url = {https://ojs.aaai.org/index.php/AAAI/article/view/10257},
	volume = {30},
	year = {2016}}

@article{testforunitroot,
 ISSN = {07350015},
 URL = {http://www.jstor.org/stable/1391432},
 author = {G. William Schwert},
 journal = {Journal of Business \& Economic Statistics},
 number = {2},
 pages = {147--159},
 publisher = {[American Statistical Association, Taylor & Francis, Ltd.]},
 title = {Tests for Unit Roots: A Monte Carlo Investigation},
 urldate = {2026-01-28},
 volume = {7},
 year = {1989}
}

@inbook{cambridge, place={Cambridge}, series={Themes in Modern Econometrics}, title={Issues in unit root testing}, booktitle={Unit Roots, Cointegration, and Structural Change}, publisher={Cambridge University Press}, author={Maddala, G. S. and Kim, In-Moo}, year={1999}, pages={98–154}, collection={Themes in Modern Econometrics}}

@ARTICLE{9514488,
  author={Zhang, Qiancheng and Ji, Hongbing and Jin, Yan},
  journal={IEEE Signal Processing Letters}, 
  title={Cyclostationary Signals Analysis Methods Based on High-Dimensional Space Transformation Under Impulsive Noise}, 
  year={2021},
  volume={28},
  number={},
  pages={1724-1728},
  doi={10.1109/LSP.2021.3104996}}

@proceedings{cyclo_a,
    author = {Stajuda, Mateusz and Garcia Cava, David and Liśkiewicz, Grzegorz},
    title = {Cyclostationary Approach for Instabilities Detection and Condition Monitoring of Centrifugal Compressor},
    volume = {Volume 2E: Turbomachinery},
    series = {Turbo Expo},
    pages = {V02ET41A032},
    year = {2020},
    month = {09},
    doi = {10.1115/GT2020-15699},
    url = {https://doi.org/10.1115/GT2020-15699},
    eprint = {https://asmedigitalcollection.asme.org/GT/proceedings-pdf/GT2020/84102/V02ET41A032/6615011/v02et41a032-gt2020-15699.pdf},
}

@article{ood_ts,
author = {Lu, Wang and Wang, Jindong and Sun, Xinwei and Chen, Yiqiang and Ji, Xiangyang and Yang, Qiang and Xie, Xing},
title = {Diversify: A General Framework for Time Series Out-of-Distribution Detection and Generalization},
year = {2024},
issue_date = {June 2024},
publisher = {IEEE Computer Society},
address = {USA},
volume = {46},
number = {6},
issn = {0162-8828},
url = {https://doi.org/10.1109/TPAMI.2024.3355212},
doi = {10.1109/TPAMI.2024.3355212},
journal = {IEEE Trans. Pattern Anal. Mach. Intell.},
month = jun,
pages = {4534–4550},
numpages = {17}
}

@inproceedings{survey_tsfm,
author = {Liang, Yuxuan and Wen, Haomin and Nie, Yuqi and Jiang, Yushan and Jin, Ming and Song, Dongjin and Pan, Shirui and Wen, Qingsong},
title = {Foundation Models for Time Series Analysis: A Tutorial and Survey},
year = {2024},
isbn = {9798400704901},
publisher = {Association for Computing Machinery},
address = {New York, NY, USA},
url = {https://doi.org/10.1145/3637528.3671451},
doi = {10.1145/3637528.3671451},
booktitle = {Proceedings of the 30th ACM SIGKDD Conference on Knowledge Discovery and Data Mining},
pages = {6555–6565},
numpages = {11},
location = {Barcelona, Spain},
series = {KDD '24}
}

@article{WEHNER2025111395,
	author = {Nikolas Wehner and Pedro Casas and Katharina Dietz and Stefan Gei{\ss}ler and Tobias Ho{\ss}feld and Michael Seufert},
	doi = {https://doi.org/10.1016/j.comnet.2025.111395},
	issn = {1389-1286},
	journal = {Computer Networks},
	pages = {111395},
	title = {Exploring the application of Time Series Foundation Models to network monitoring tasks},
	url = {https://www.sciencedirect.com/science/article/pii/S1389128625003627},
	volume = {269},
	year = {2025}}

@INPROCEEDINGS{indus_pc,
  author={Yan, Peng and Abdulkadir, Ahmed and Schatte, Gerrit A. and Aguzzi, Giulia and Gha, Joonsu and Pascher, Nikola and Rosenthal, Matthias and Gao, Yunlong and Grewe, Benjamin F. and Stadelmann, Thilo},
  booktitle={2025 IEEE Swiss Conference on Data Science (SDS)}, 
  title={Learning Actionable World Models for Industrial Process Control}, 
  year={2025},
  volume={},
  number={},
  pages={111-118},
  doi={10.1109/SDS66131.2025.00022}}

@article{Benneyan458,
	author = {Benneyan, J C and Lloyd, R C and Plsek, P E},
	doi = {10.1136/qhc.12.6.458},
	eprint = {https://qualitysafety.bmj.com/content/12/6/458.full.pdf},
	issn = {1475-3898},
	journal = {BMJ Quality \& Safety},
	number = {6},
	pages = {458--464},
	publisher = {BMJ Publishing Group Ltd},
	title = {Statistical process control as a tool for research and healthcare improvement},
	url = {https://qualitysafety.bmj.com/content/12/6/458},
	volume = {12},
	year = {2003}}

\newpage
\appendix
\onecolumn
\section{Preliminaries}

\subsection{Stationarity and Unit Root}

Let $\{X_t\}_{t \in \mathbb{Z}}$ denote a real-valued discrete-time stochastic
process.
Throughout this paper, we focus on univariate time series, consistent with the
design philosophy of the underlying time-series foundation model considered in
this work.

The process $\{X_t\}$ is said to be \emph{weakly stationary} if it satisfies the
following conditions:
\begin{enumerate}
    \item $\mathbb{E}[X_t^2] < \infty$, for all $t \in \mathbb{Z}$;
    \item $\mathbb{E}[X_t] = \mu$, for all $t \in \mathbb{Z}$;
    \item $\gamma_X(s,t) = \gamma_X(s+h,t+h)$, for all $s,t,h \in \mathbb{Z}$,
\end{enumerate}
where $\gamma_X(s,t) = \mathrm{Cov}(X_s, X_t)$ denotes the autocovariance function\cite{Cryer_Chan_2008}.
Equivalently, a weakly stationary process has finite second moments, a constant
first moment, and second-order statistics that depend only on the time lag
$|t-s|$, and not on the absolute time indices.
Unless stated otherwise, the term \emph{stationarity} refers to weak stationarity.
When stationarity in the strict sense is intended, we explicitly use the term
\emph{strict stationarity}.

A canonical violation of weak stationarity arises from the presence of a
\emph{unit root}.
Consider an autoregressive process of order one,
\begin{equation}
    X_t = \phi X_{t-1} + \varepsilon_t,
\end{equation}
where $\{\varepsilon_t\}$ is a zero-mean white noise process with finite variance.
When $|\phi| < 1$, the process is weakly stationary.
In contrast, when $\phi = 1$, the process reduces to a random walk,
\begin{equation}
    X_t = X_{t-1} + \varepsilon_t,
\end{equation}
for which the variance grows unbounded over time.
This violates the finite-variance requirement of weak stationarity and leads to
time-dependent second-order statistics.
Unit root behavior is therefore widely regarded as a representative and
theoretically well-studied form of non-stationarity.

\subsection{Cyclostationarity}

Classical definitions of stationarity require statistical properties to be
time-invariant.
However, many real-world signals exhibit structured temporal variation that
violates this requirement while remaining highly regular.
A stochastic process $\{X_t\}$ is said to be \emph{cyclostationary} with period
$T$ if its statistics vary periodically over time.
In particular, a cyclostationary process may have a time-dependent but periodic
mean,
\begin{equation}
    \mathbb{E}[X_t] = \mu(t), \quad \mu(t+T) = \mu(t),
\end{equation}
and a periodically varying autocovariance,
\begin{equation}
    \gamma_X(t,h) = \mathrm{Cov}(X_t, X_{t+h}), \quad
    \gamma_X(t+T,h) = \gamma_X(t,h).
\end{equation}
Thus, unlike weakly stationary processes, cyclostationary processes do not
require constant first- or second-order moments, but instead exhibit stability
across cycles.

Cyclostationarity has been extensively studied in signal processing and
communications, where periodic structure naturally arises from modulation,
synchronization, or seasonal effects.
Dedicated cyclostationarity tests explicitly exploit this periodic structure,
for example by analyzing cyclic frequencies or by comparing statistics across
corresponding phases of the cycle.
These methods are particularly effective when the underlying periodicity is
strong and well-defined.

A prominent class of cyclostationarity tests is based on generalized likelihood
ratio testing (GLRT), which compares likelihoods under stationary and
cyclostationary hypotheses by explicitly modeling periodic second-order
statistics.
Such tests are typically formulated as binary hypothesis tests and often rely on
assumptions regarding the period or cyclic structure.
While powerful in their intended setting, they are not designed to capture more
general forms of non-stationarity, such as abrupt regime changes, variance
shifts, or mixed deviations that do not conform to a single periodic pattern.

\subsection{Cyclostationarity Test of Dandawate and Giannakis}
\label{app:dg_test}

The cyclostationarity test proposed by Dandawate and Giannakis~\cite{DG_TEST}
is a frequency-domain method designed to detect second-order cyclostationary
structure in time series.
The test is based on estimating cyclic autocorrelations at a specified cyclic
frequency and constructing a quadratic form whose asymptotic distribution
under the null hypothesis of no cyclostationarity follows a $\chi^2$ law.

In our experiments, the DG test is used as a secondary ``rescue'' test following
a first-stage stationarity test. In our implementation, the cyclic frequency is not assumed to be known a priori.
Instead, for each window, we estimate a dominant frequency via the periodogram
and apply the DG test at the estimated frequency.
To ensure resolvability, the estimated frequency is required to exceed a
minimum threshold of $2/L$, where $L$ denotes the window length.

Given the estimated cyclic frequency, we compute the DG test statistic and
compare it against the corresponding $\chi^2$ critical value at the specified
significance level.
If the statistic exceeds the threshold, the window is classified as exhibiting
cyclostationary structure and is reclassified as in-regime in the two-stage
baseline.
This implementation avoids oracle assumptions and is applied uniformly across
all DG-based baselines, ensuring a fair and reproducible comparison.

\section{Additional Experiments}
\subsection{Fixed Data Length Setting}
\label{app:fixed}

While the main experiments focus on mixed window lengths to reflect realistic
monitoring scenarios, we additionally evaluate the proposed framework under a
fixed-length setting.
This experiment isolates the effect of window-length heterogeneity and serves
as a controlled sanity check for both detection performance and calibration.

\paragraph{Setup.}
All windows—reference, calibration, and test—are generated with a fixed length
$L$.
We consider the same data-generating processes as in the main experiments,
including stationary AR(1) regimes and stable cyclostationary regimes, together
with the corresponding deviation classes.
The reference set, calibration set, and test set are constructed following the
same reference-relative protocol described in Section~\ref{sec:experiments},
but without length variation.

All experiments in this section use the Chronos2 encoder with mean pooling,
identical projection choices, and the same conformal calibration procedure as
in the mixed-length setting.
No per-length conditioning is required here, since exchangeability holds
trivially under a fixed window length.
\begin{table}[t]
\centering
\caption{\textbf{Fixed-length setting (AR(1) reference regime).} Window lengths are fixed per run and results are aggregated across lengths (mean $\pm$ std over 8 runs: 8 lengths $\times$ 1 seed in this ablation).}
\label{tab:fixed_ar1}
\resizebox{0.95\columnwidth}{!}{
\begin{tabular}{l|cc|ccc}
\toprule
\textbf{Method} & \multicolumn{2}{c|}{\textbf{Discrimination}} & \multicolumn{3}{c}{\textbf{Detection ($\alpha=0.05$)}} \\
& ROC-AUC & PR-AUC & F1-Score & Precision & Recall \\
\midrule
\textbf{Proposed} & \textbf{0.952 $\pm$ 0.000} & \textbf{0.967 $\pm$ 0.001} & \textbf{0.924 $\pm$ 0.001} & \textbf{0.961 $\pm$ 0.001} & \textbf{0.890 $\pm$ 0.001} \\
KPSS & 0.770 $\pm$ 0.002 & 0.765 $\pm$ 0.004 & 0.702 $\pm$ 0.005 & 0.795 $\pm$ 0.006 & 0.628 $\pm$ 0.006 \\
ADF & 0.742 $\pm$ 0.009 & 0.753 $\pm$ 0.013 & 0.686 $\pm$ 0.009 & 0.709 $\pm$ 0.009 & 0.664 $\pm$ 0.010 \\
PP & 0.741 $\pm$ 0.008 & 0.754 $\pm$ 0.013 & 0.684 $\pm$ 0.009 & 0.707 $\pm$ 0.010 & 0.663 $\pm$ 0.011 \\
\bottomrule
\end{tabular}
}
\end{table}

\begin{table}[t]
\centering
\caption{\textbf{Fixed-length setting (combined stationary--cyclostationary reference regime).} Window lengths are fixed per run and results are aggregated across lengths (mean $\pm$ std over 8 runs).}
\label{tab:fixed_view2}
\resizebox{0.95\columnwidth}{!}{
\begin{tabular}{l|cc|ccc}
\toprule
\textbf{Method} & \multicolumn{2}{c|}{\textbf{Discrimination}} & \multicolumn{3}{c}{\textbf{Detection ($\alpha=0.05$)}} \\
& ROC-AUC & PR-AUC & F1-Score & Precision & Recall \\
\midrule
\textbf{Proposed} & \textbf{0.952 $\pm$ 0.000} & \textbf{0.967 $\pm$ 0.001} & \textbf{0.924 $\pm$ 0.001} & \textbf{0.961 $\pm$ 0.001} & \textbf{0.890 $\pm$ 0.001} \\
ClaSP & 0.658 $\pm$ 0.010 & 0.650 $\pm$ 0.014 & 0.478 $\pm$ 0.015 & 0.695 $\pm$ 0.020 & 0.364 $\pm$ 0.015 \\
CUSUM & 0.609 $\pm$ 0.004 & 0.613 $\pm$ 0.006 & 0.497 $\pm$ 0.007 & 0.598 $\pm$ 0.009 & 0.425 $\pm$ 0.007 \\
ADF+DG & 0.537 $\pm$ 0.004 & 0.542 $\pm$ 0.005 & 0.482 $\pm$ 0.005 & 0.499 $\pm$ 0.006 & 0.466 $\pm$ 0.004 \\
KPSS+DG & 0.524 $\pm$ 0.005 & 0.530 $\pm$ 0.006 & 0.484 $\pm$ 0.006 & 0.499 $\pm$ 0.007 & 0.469 $\pm$ 0.006 \\
PP+DG & 0.536 $\pm$ 0.004 & 0.541 $\pm$ 0.006 & 0.482 $\pm$ 0.005 & 0.499 $\pm$ 0.006 & 0.466 $\pm$ 0.005 \\
\bottomrule
\end{tabular}
}
\end{table}
\vspace{1.0em}

\paragraph{Results.}
Tables~\ref{tab:fixed_ar1} and~\ref{tab:fixed_view2} report performance under
fixed-length windows for the AR(1) reference regime and the combined
stationary--cyclostationary reference regime, respectively.
Overall trends closely mirror those observed in the mixed-length setting.
The proposed method achieves consistently high AUC across all deviation types,
substantially outperforming classical stationarity tests and change-point
detection baselines.

Notably, empirical false-alarm rates under conformal calibration are well
aligned with the nominal level $\alpha$, in contrast to the global calibration
behavior observed when pooling across heterogeneous lengths.
This confirms that the miscalibration observed in the mixed-length setting
stems from length heterogeneity rather than from the conformal procedure
itself.

\paragraph{Discussion.}
These results highlight two key points.
First, under fixed window lengths, standard conformal calibration is sufficient
to ensure valid type-I error control, and no conditional or stratified variant
is required.
Second, the strong performance of the proposed framework in this simplified
setting confirms that its advantages are not an artifact of length mixing or
calibration heuristics.

Taken together with the mixed-length experiments in the main paper, this
appendix clarifies that length-conditional calibration is not a modeling choice
but a structural necessity induced by heterogeneous window lengths.

\begin{table}[t]
\centering
\caption{
Projection ablation study using the Chronos2 encoder.
We compare supervised LDA ($d=1$) with unsupervised PCA and random projections
under mixed window lengths.
Performance is reported using standard detection metrics
(ROC-AUC, PR-AUC, F1-score, precision, and recall),
with decision thresholds determined via conformal calibration at $\alpha=0.05$.
}
\label{tab:proj_ch2}
\resizebox{0.95\columnwidth}{!}{
\begin{tabular}{ll|cc|ccc}
\toprule
\multicolumn{2}{c|}{Method} & \multicolumn{2}{c|}{Discrimination} & \multicolumn{3}{c}{Detection ($\alpha=0.05$)} \\
Type & Dim ($d$) & ROC-AUC & PR-AUC & F1-Score & Precision & Recall \\
\midrule
\textbf{LDA (Proposed)} & \textbf{1}
& \textbf{0.952} & \textbf{0.967}
& \textbf{0.924} & \textbf{0.961} & \textbf{0.890} \\
\midrule
\multirow{4}{*}{PCA}
 & 1 & 0.842 & 0.862 & 0.688 & 0.919 & 0.550 \\
 & 2 & 0.868 & 0.890 & 0.727 & 0.927 & 0.598 \\
 & 4 & 0.862 & 0.890 & 0.727 & 0.927 & 0.598 \\
 & 8 & 0.871 & 0.872 & 0.709 & 0.918 & 0.578 \\
\midrule
\multirow{3}{*}{Random}
 & 1 & 0.607 & 0.619 & 0.245 & 0.748 & 0.147 \\
 & 4 & 0.770 & 0.730 & 0.360 & 0.808 & 0.232 \\
 & 8 & 0.843 & 0.821 & 0.533 & 0.883 & 0.381 \\
\bottomrule
\end{tabular}
}
\end{table}

\begin{table}[t]
\centering
\caption{Projection ablation study (MOMENT).
We compare supervised LDA with unsupervised PCA and random projections under mixed sequence lengths.
Detection metrics (ROC-AUC, PR-AUC, F1, precision, and recall) are computed from window-level decisions,
with decision thresholds determined via conformal calibration at $\alpha=0.05$.
}
\label{tab:proj_mom}
\begin{tabular}{l c|cccccc}
\toprule
Method & Dim 
& ROC-AUC 
& PR-AUC 
& F1 
& Precision 
& Recall \\
\midrule
\textbf{LDA} & 1 
& 0.8598
& 0.9001 
& 0.7721 
& 0.9557 
& 0.6477 \\
\midrule
PCA & 1 
& 0.3819 
& 0.4081 
& 0.0000 
& 0.0000 
& 0.0000 \\
PCA & 2 
& 0.3598 
& 0.3984 
& 0.0000 
& 0.0000 
& 0.0000 \\
PCA & 4 
& 0.5104 
& 0.5504 
& 0.0000 
& 0.0000 
& 0.0000 \\
PCA & 8 
& 0.5697 
& 0.6204 
& 0.0063 
& 0.8750 
& 0.0032 \\
\midrule
Random & 1 
& 0.5097 
& 0.5062 
& 0.0345 
& 0.6290 
& 0.0177 \\
Random & 2 
& 0.3836 
& 0.4237 
& 0.0700 
& 0.3346 
& 0.0391 \\
Random & 4 
& 0.4907 
& 0.4954 
& 0.0351 
& 0.5063 
& 0.0182 \\
Random & 8 
& 0.5527 
& 0.5559 
& 0.0747 
& 0.6692 
& 0.0395 \\
\bottomrule
\end{tabular}
\end{table}

\subsection{Choice of Encoder Model}
\label{app:encoder}

The proposed framework relies on pretrained time-series encoders to map local windows
into a representation space where distributional consistency can be assessed.
To examine the sensitivity of the method to the choice of encoder, we compare three
representative time-series foundation models: Chronos2, MOMENT, and TOTEM\cite{chronos2, moment, talukder2024totem}.

Table~\ref{tab:encoder_ablation_view1} and Table~\ref{tab:encoder_ablation_view2}
summarize performance under View~1 (stationary and cyclostationary reference regimes)
and View~2 (cyclostationary deviation settings), respectively.
Chronos2 consistently achieves the strongest performance across all metrics,
indicating that its representations preserve fine-grained temporal structure
relevant for reference-relative monitoring.
MOMENT performs comparably under View~1 and degrades gracefully under View~2,
suggesting partial sensitivity to cyclostationary deviations.
In contrast, TOTEM fails to produce discriminative representations in both views,
often performing near or below chance level.

These results indicate that while the proposed framework is modular and encoder-agnostic
in principle, practical performance depends on the quality of the learned representation.
In all experiments, encoders are used as frozen feature extractors without task-specific
fine-tuning.

\subsection{Choice of Projection Method}
\label{app:proj}
Tables~\ref{tab:proj_ch2} and~\ref{tab:proj_mom}
compare different projection strategies, including random projection, PCA, and
supervised LDA, under identical experimental settings.
Across both encoders, the choice of projection has a non-negligible impact on
detection performance as measured by ROC-AUC, PR-AUC, F1 score, precision, and recall.

For the Chronos2 encoder, supervised LDA yields consistently higher ROC-AUC and
PR-AUC across most settings, with particularly pronounced gains in mixed-length
regimes.
Unsupervised PCA provides moderate improvements over random projection, while random
projections show less consistent behavior.
These results suggest that emphasizing directions sensitive to known distributional
deviations can improve the statistical efficiency of subsequent density-based scoring,
even though the projection is learned independently of the reference and test sets.

\begin{table}[t]
\centering
\caption{
Encoder ablation results under View~1 (stationary / cyclostationary reference regimes).
Chronos2 and MOMENT produce highly discriminative representations, while TOTEM fails
to separate reference and out-of-regime windows.
}
\label{tab:encoder_ablation_view1}
\begin{tabular}{lccccc}
\toprule
Encoder & ROC-AUC & PR-AUC & F1 & Precision & Recall \\
\midrule
Chronos2 & 0.9974 & 0.9961 & 0.9664 & 0.9421 & 0.9920 \\
MOMENT   & 0.9968 & 0.9974 & 0.9660 & 0.9515 & 0.9810 \\
TOTEM    & 0.0458 & 0.3176 & 0.4887 & 0.3929 & 0.6460 \\
\bottomrule
\end{tabular}
\end{table}

\begin{table}[t]
\centering
\caption{
Encoder ablation results under View~2 (cyclostationary deviation settings).
Chronos2 remains robust, while MOMENT degrades gracefully and TOTEM performs poorly.
}
\label{tab:encoder_ablation_view2}
\begin{tabular}{lccccc}
\toprule
Encoder & ROC-AUC & PR-AUC & F1 & Precision & Recall \\
\midrule
Chronos2 & 0.9859 & 0.9827 & 0.9404 & 0.9478 & 0.9332 \\
MOMENT   & 0.8557 & 0.8836 & 0.7013 & 0.9342 & 0.5614 \\
TOTEM    & 0.3564 & 0.3957 & 0.1755 & 0.2477 & 0.1359 \\
\bottomrule
\end{tabular}
\end{table}

\begin{table}[h]
\centering
\caption{Sensitivity to the number of reference windows ($n_{\text{ref}}$).
Performance is reported under mixed sequence lengths using conformal calibration at $\alpha=0.05$. by MOMENT}
\label{tab:nref_sensitivity}
\begin{tabular}{c|cccccc}
\toprule
$n_{\text{ref}}$ 
& ROC-AUC 
& PR-AUC 
& $\tau$ ($\alpha=0.05$)
& F1 
& Precision 
& Recall \\
\midrule
5{,}000  
& 0.8457 
& 0.8877 
& 2.7547
& 0.7748 
& 0.9578 
& 0.6505 \\
10{,}000  
& 0.8598 
& 0.9001 
& 2.8033
& 0.7721 
& 0.9557 
& 0.6477 \\
15{,}000  
& 0.8467 
& 0.8898 
& 2.7968
& 0.7758 
& 0.9542 
& 0.6536 \\
20{,}000  
& 0.8509 
& 0.8956 
& 2.8516
& 0.7727 
& 0.9545 
& 0.6491 \\
25{,}000  
& 0.8369 
& 0.8876 
& 2.8951
& 0.7652 
& 0.9583 
& 0.6368 \\
50{,}000  
& 0.8410 
& 0.8875 
& 2.8802
& 0.7579 
& 0.9532 
& 0.6291 \\
75{,}000  
& 0.8403 
& 0.8878 
& 2.8655
& 0.7819 
& 0.9542 
& 0.6623 \\
100{,}000 
& 0.8366 
& 0.8816 
& 2.8227
& 0.7718 
& 0.9441 
& 0.6527 \\
\bottomrule
\end{tabular}
\end{table}

\begin{table*}[t]
\centering
\caption{Overall detection performance comparison across different encoders and sequence lengths. \textbf{Chronos2} demonstrates the best performance, peaking at sequence length 336. \textbf{TOTEM} fails to extract meaningful anomaly features.}
\label{tab:ablation_encoder_seq}
\resizebox{\textwidth}{!}{
\begin{tabular}{l|ccc|ccc|ccc}
\toprule
 & \multicolumn{3}{c|}{\textbf{Chronos2 (Proposed)}} & \multicolumn{3}{c|}{\textbf{MOMENT}} & \multicolumn{3}{c}{\textbf{TOTEM}} \\
Seq Len & ROC-AUC & F1-Score & Recall & ROC-AUC & F1-Score & Recall & ROC-AUC & F1-Score & Recall \\
\midrule
64  & 0.952 & 0.905 & 0.857 & 0.908 & 0.826 & 0.755 & 0.452 & 0.437 & 0.415 \\
96  & 0.965 & 0.923 & 0.886 & 0.940 & 0.865 & 0.795 & 0.478 & 0.530 & 0.535 \\
128 & 0.969 & 0.930 & 0.897 & 0.955 & 0.899 & 0.860 & 0.490 & 0.538 & 0.546 \\
192 & 0.972 & 0.929 & 0.895 & 0.960 & \textbf{0.910} & 0.855 & 0.465 & 0.584 & 0.612 \\
256 & 0.974 & 0.937 & 0.904 & 0.963 & 0.907 & \textbf{0.862} & 0.474 & 0.589 & 0.619 \\
336 & \textbf{0.977} & \textbf{0.941} & 0.909 & \textbf{0.967} & 0.908 & 0.856 & 0.471 & 0.591 & 0.620 \\
512 & 0.973 & 0.938 & 0.908 & 0.962 & 0.881 & 0.808 & 0.479 & 0.603 & 0.640 \\
720 & 0.975 & 0.941 & \textbf{0.910} & 0.964 & 0.891 & 0.827 & 0.480 & \textbf{0.625} & \textbf{0.674} \\
\bottomrule
\end{tabular}
}
\end{table*}

\label{app:data_generation}
\begin{table}[t]
\centering
\caption{
Parameter ranges used for synthetic data generation.
All parameters are independently sampled per window unless stated otherwise.
Ranges are chosen to induce diverse but well-controlled stationary and
cyclostationary regimes, as well as structured deviations within individual windows.
}
\label{tab:data_params}
\resizebox{0.95\columnwidth}{!}{
\begin{tabular}{l|l|l}
\toprule
\textbf{Component} & \textbf{Parameter} & \textbf{Range / Description} \\
\midrule
\multicolumn{3}{c}{\textbf{AR(1) Reference Regime}} \\
\midrule
AR(1) & Autoregressive coefficient $\phi$ & $\mathrm{Unif}(-0.9,\,0.9)$ \\
 & Mean $\mu$ & $\mathrm{Unif}(-0.5,\,0.5)$ \\
 & Noise std.\ $\sigma$ & $\mathrm{Unif}(0.02,\,0.30)$ \\
\midrule
\multicolumn{3}{c}{\textbf{AR(1) Deviations}} \\
\midrule
Mean shift & Shift magnitude & $\mathrm{Unif}(0.2,\,0.6)$ (random sign) \\
Variance shift & $\sigma_1, \sigma_2$ & $\mathrm{Unif}(0.03,\,0.06)$ / $\mathrm{Unif}(0.12,\,0.20)$ \\
Trend & Slope & $\mathrm{Unif}(0.3,\,0.6)$ (random sign) \\
Unit-root & Noise std.\ $\sigma$ & $\mathrm{Unif}(0.05,\,0.20)$ \\
\midrule
\multicolumn{3}{c}{\textbf{Periodic-Template Reference Regime}} \\
\midrule
Template & Period divisor $k$ & $\{1,2,4,8\}$ \\
 & Period $T$ & $T = L/k$ \\
 & Phase offset $\phi$ & $\mathrm{Unif}\{0,\dots,T-1\}$ \\
 & Amplitude $A$ & $\mathrm{Unif}(0.5,\,1.5)$ \\
 & Noise std.\ $\sigma$ & $\mathrm{Unif}(0.02,\,0.10)$ \\
\midrule
\multicolumn{3}{c}{\textbf{Periodic Deviations}} \\
\midrule
Amplitude step & Step magnitude & $\mathrm{Unif}(0.3,\,0.8)$ (random sign) \\
Amplitude drift & Drift slope & $\mathrm{Unif}(0.01,\,0.04)$ (random sign) \\
Template distortion & Perturbation & Additive noise on template shape \\
Frequency mismatch & Frequency change & Swap $k$ within $\{1,2,4,8\}$ \\
Compound deviations & Combination & Amplitude step + drift \\
Compound-structural deviations & Combination & Frequency mismatch + envelope/template perturbation \\
\bottomrule
\end{tabular}
}
\end{table}

\section{Implementation Details}
\label{app:imp_detail}

\paragraph{GPU Setup}
All experiments were conducted using a single NVIDIA RTX 3090 GPU with 24GB memory.

\paragraph{Window construction.}
We use mixed window lengths drawn from \{64, 96, 128, 192, 256, 336, 512, 720\}, reflecting both common practice in time series modeling and multi-scale temporal coverage. In addition, we did fixed length setting which assume that input size is fixed. Results are can be seen in Appendix \ref{app:fixed}

\begin{figure*}[t]
\centering

% ---------- Top figure ----------
\begin{minipage}[t]{0.95\textwidth}
    \centering
    \includegraphics[width=\linewidth]{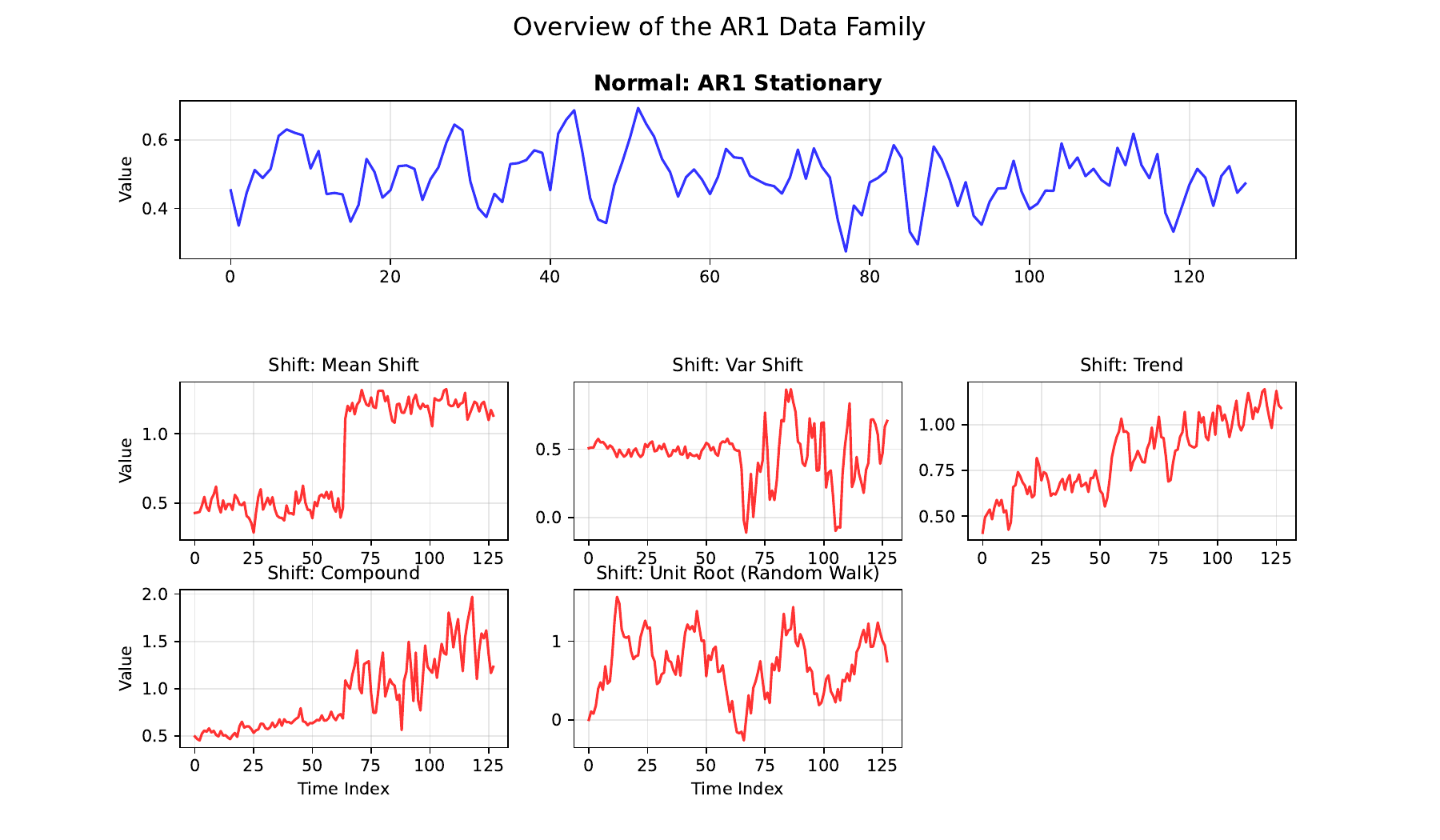}
    \captionof{figure}{
    AR(1) data family.
    Top: stationary AR(1) windows.
    Bottom: representative deviations, including mean shift, variance shift,
    linear trend, unit-root behavior, and combined changes.
    }
    \label{fig:ar1_family}
\end{minipage}

\vspace{1.2em}  % <-- controls vertical balance

% ---------- Bottom figure ----------
\begin{minipage}[t]{0.95\textwidth}
    \centering
    \includegraphics[width=\linewidth]{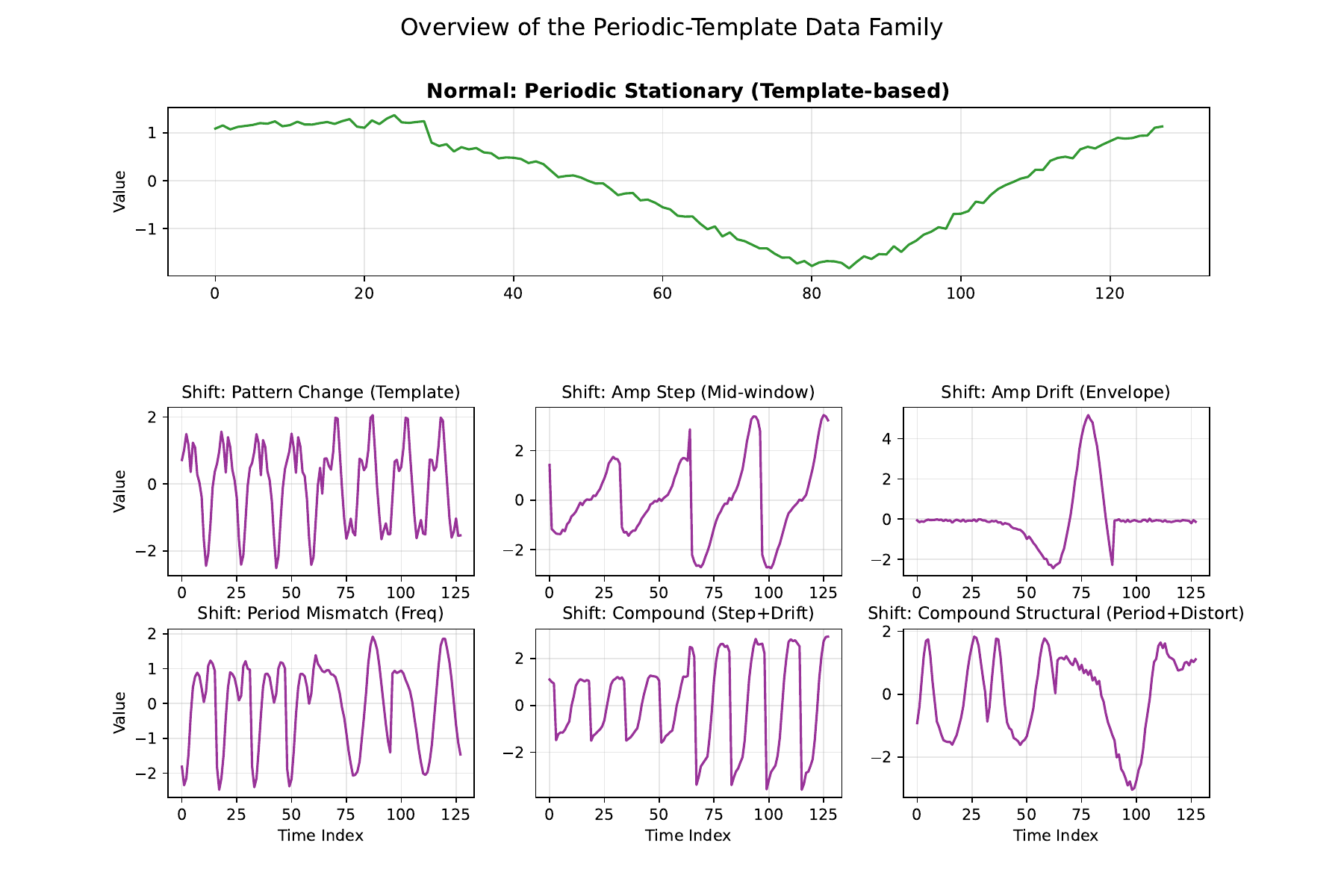}
    \captionof{figure}{
    Periodic-template data family.
    Top: stable periodic regime generated by repeating an unknown template.
    Bottom: representative deviations including amplitude, frequency,
    and template perturbations.
    }
    \label{fig:periodic_family}
\end{minipage}

\end{figure*}

\paragraph{Deviations from AR(1) stationarity.}
To evaluate sensitivity to structured departures from stationarity,
we introduce several types of within-window deviations:
(i) \emph{mean shifts}, where the mean changes halfway through the window;
(ii) \emph{variance shifts}, where the innovation variance changes;
(iii) \emph{linear trends}, where a deterministic slope is added;
(iv) \emph{combined shifts}, involving multiple simultaneous changes;
and (v) \emph{unit-root dynamics}, implemented as random-walk behavior.
Representative examples are shown in Fig.~\ref{fig:ar1_family}.
These deviations preserve local temporal structure while violating
distributional consistency with the reference regime.

\paragraph{Stable periodic-template regime.}
To model a stable periodic operating regime without restricting the waveform to a sinusoid, we generate windows by repeating an unknown template.
For a fixed window length $L$, we sample
$
k \sim \mathrm{Unif}\{1,2,4,8\}
$
and set the period as
$
T = L/k
$
(assuming $k$ divides $L$).
Let $g_T \in \mathbb{R}^{T}$ denote a template sequence, normalized to have zero mean and unit variance.
We also sample a phase offset $\phi \sim \mathrm{Unif}\{0,\dots,T-1\}$ and an amplitude $A$.
Then the window $x \in \mathbb{R}^{L}$ is generated as
\begin{equation}
x_t
=
A\, g_T\!\big((t+\phi)\bmod T\big)
+
\epsilon_t,
\quad
t=0,\dots,L-1,
\end{equation}
where $\epsilon_t \sim \mathcal{N}(0,\sigma^2)$.
This construction yields a stable periodic regime in the sense that the distribution of windows is invariant under time shifts by multiples of $T$, while allowing rich within-cycle shapes via $g_T$.

\paragraph{Deviations from periodic stability.}
We introduce several classes of deviations from the periodic-template regime:
(i) \emph{amplitude steps}, where the signal amplitude changes abruptly
within a window;
(ii) \emph{amplitude drift}, inducing gradual envelope variation;
(iii) \emph{template distortion}, where the repeated pattern itself changes;
and
(iv) \emph{frequency mismatch}, corresponding to a change in the underlying
period within a window.

In addition, we consider two composite deviation classes.
\emph{Compound deviations} combine multiple low-order perturbations
(e.g., amplitude steps and drift) while preserving a fixed underlying
periodic structure.
In contrast, \emph{compound structural deviations} introduce higher-order
changes that alter the periodic organization itself, such as frequency
mismatch combined with envelope or template perturbations.
Both classes retain strong periodic components but violate distributional
consistency with the reference regime in qualitatively different ways.

Representative examples are shown in Fig.~\ref{fig:periodic_family} and detailed parameter ranges for all synthetic data generation procedures
are provided in Table~\ref{tab:data_params}.

\paragraph{Encoders and representations.}
We evaluate multiple pretrained time-series encoders to obtain fixed-dimensional
window representations.
Unless stated otherwise, Chronos2 is used as the primary encoder in all main
experiments.

For Chronos2, we obtain embeddings by extracting hidden states from the pretrained
model and applying mean pooling across the temporal dimension.
The model is loaded directly from the HuggingFace repository\footnote{\url{https://huggingface.co/amazon/chronos-2}}, using the official
pretrained weights, and no fine-tuning is performed.
This yields a single embedding vector $f_\theta(x) \in \mathbb{R}^D$ for each
window, regardless of its length.

For MOMENT, we use the official pretrained implementation provided by the authors,
installed via \texttt{pip install momentfm}\footnote{\url{https://github.com/moment-timeseries-foundation-model/moment}}.
We use MOMENT solely as a representation extractor by discarding the task-specific
prediction head and retaining only the embedding backbone.
No task-specific fine-tuning or supervision is applied.

For TOTEM, we use the pretrained forecasting tokenizer provided in the official
repository\footnote{\url{https://github.com/SaberaTalukder/TOTEM}}.
The tokenizer maps each input window into a latent representation space learned
via a forecasting objective.
As with other encoders, we treat TOTEM as a frozen feature extractor and do not
apply any task-specific fine-tuning.

Across all encoders, representations are computed independently for each window
and are used solely as inputs to the subsequent projection, density modeling,
and calibration stages.

\paragraph{Baselines and reproducibility.}
Classical stationarity tests and change-point detection baselines are implemented using standard configurations. All results are averaged over multiple runs.

\begin{table}[h]
\centering
\caption{Implementation details and experimental configuration checklist.}
\label{tab:impl_checklist}
\begin{tabular}{p{4.2cm} p{8.2cm}}
\toprule
\textbf{Component} & \textbf{Configuration} \\
\midrule
Window length ($\texttt{seq\_len}$) 
& Mixed lengths drawn from $\{64, 96, 128, 192, 256, 336, 512, 720\}$, covering short- to long-range temporal contexts. \\

Window length sampling 
& Discrete mixture over predefined lengths to reflect heterogeneous monitoring conditions. \\

Reference regime 
& Stationary or cyclostationary windows only; no anomalous samples included. \\

Representation model 
& Pretrained time-series foundation models used as frozen encoders (Chronos2\cite{chronos2} as primary backbone). \\

Fine-tuning 
& None. Encoders are used without task-specific training. \\

Projection method 
& Linear Discriminant Analysis (LDA) \\

Projection dimension 
& $d = 1$ (default); higher dimensions evaluated in ablation studies. \\

Score construction 
& Negative log-density under a Gaussian kernel density estimator (KDE) fitted on projected reference representations. \\

Kernel and bandwidth 
& Gaussian kernel with bandwidth selected automatically via Silverman's rule of thumb; no tuning on test data. \\

Calibration method 
& Conformal calibration applied to reference scores to control false alarm rates. \\

Decision threshold 
& Implicitly determined by target false alarm rate via conformal calibration; no manual tuning. \\

Baseline methods 
& Classical stationarity tests (ADF, KPSS, PP) applied at the window level using standard configurations. \\

Change-point detection 
& Evaluated separately using off-the-shelf implementations; not included in main process control tables. \\

Sequence length mismatch 
& Reference and test windows may have different lengths; representations are length-invariant. \\

Hyperparameter tuning 
& No tuning performed on test data; default settings used unless otherwise specified. \\

Randomness control 
& Experiments repeated over multiple random seeds; mean and standard deviation reported. \\

Label usage 
& Labels used only for evaluation; not used in reference modeling, projection, or calibration. \\
\bottomrule
\end{tabular}
\end{table}

\begin{algorithm}[tb]
\caption{Reference-Relative Consistency Testing (Length-Conditional Conformal)}
\label{alg:conformal}
\begin{algorithmic}

\STATE {\bfseries Input:}
\STATE \hspace{1em} Reference windows $\{x_i \in \mathbb{R}^{L_i}\}_{i=1}^N$
\STATE \hspace{1em} Test window $x_{\text{test}} \in \mathbb{R}^{L_{\text{test}}}$
\STATE \hspace{1em} Encoder $f_\theta:\mathbb{R}^L\!\to\!\mathbb{R}^D$, projection $\Pi:\mathbb{R}^D\!\to\!\mathbb{R}^d$
\STATE \hspace{1em} Significance level $\alpha$

\vspace{0.6em}
\STATE {\bfseries 1. Reference Modeling}
\STATE \hspace{1em} Split reference into $\mathcal{R}_{\text{fit}}$ and $\mathcal{R}_{\text{cal}}$
\STATE \hspace{1em} Fit density $\hat p(\cdot)$ on $\{\Pi(f_\theta(x)) : x \in \mathcal{R}_{\text{fit}}\}$

\vspace{0.6em}
\STATE {\bfseries 2. Scoring}
\STATE \hspace{1em} $s(x) \leftarrow -\log \hat p(\Pi(f_\theta(x)))$
\STATE \hspace{1em} Compute $s_{\text{test}}$ and calibration scores $\{s_i\}$

\vspace{0.6em}
\STATE {\bfseries 3. Length-Conditional Calibration}
\STATE \hspace{1em} Select calibration set $\mathcal{R}_{\text{cal}}^{(L_{\text{test}})}$
\STATE \hspace{1em} Compute
\[
p_{\text{conf}} =
\frac{
1 + \sum_{x_i \in \mathcal{R}_{\text{cal}}^{(L_{\text{test}})}} 
\mathbb{I}[s_i \ge s_{\text{test}}]
}{
|\mathcal{R}_{\text{cal}}^{(L_{\text{test}})}| + 1
}
\]

\vspace{0.4em}
\STATE {\bfseries Output:}
\STATE \hspace{1em} $\delta(x_{\text{test}})=\mathbb{I}[p_{\text{conf}}>\alpha]$

\end{algorithmic}
\end{algorithm}

\end{document}